\documentclass[12pt,a4paper]{article}                                    
\usepackage{a4wide}

\usepackage{graphics} 
\usepackage{epsfig} 
\usepackage{amsmath} 

\begin{document}

\title{Distance estimation with efference copies and optical flow maneuvers: a stability-based strategy.} 
\author{Guido C.H.E. de Croon\thanks{\texttt g.c.h.e.decroon@tudelft.nl}}
\date{}

\maketitle

\begin{abstract}

The visual cue of optical flow plays a major role in the navigation of flying insects, and is increasingly studied for use by small flying robots as well. A major problem is that successful optical flow control seems to require distance estimates, while optical flow is known to provide only the ratio of velocity to distance. In this article, a novel, stability-based strategy is proposed to estimate distances with monocular optical flow and knowledge of the control inputs (efference copies). It is shown analytically that given a fixed control gain, the stability of a constant divergence control loop only depends on the distance to the approached surface. At close distances, the control loop first starts to exhibit self-induced oscillations, eventually leading to instability. The proposed stability-based strategy for estimating distances has two major attractive characteristics. First, self-induced oscillations are easy for the robot to detect and are hardly influenced by wind. Second, the distance can be estimated during a zero divergence maneuver, i.e., around hover. The stability-based strategy is implemented and tested both in simulation and with a Parrot AR drone $2.0$. It is shown that it can be used to: (1) trigger a final approach response during a constant divergence landing with fixed gain, (2) estimate the distance in hover, and (3) estimate distances during an entire landing if the robot uses adaptive gain control to continuously stay on the `edge of oscillation'.

\end{abstract}


\section{Introduction}
A major challenge in robotics is to achieve autonomous operation of tiny flying robots such as 25g ``pocket drones'' \cite{DUNKLEY2014} or more extremely the 80mg Robobee \cite{MA2013}. Flying insects provide a rich source of inspiration for solving this challenge, since they are able to navigate successfully on the basis of a very limited sensory and processing apparatus. Flying insects rely heavily on optical flow, i.e., the apparent motion of world points caused by the relative motion between the observer and the environment \cite{GIBSON1979}. Also for robots this cue is promising as it can be extracted from a single passive camera, implying light weight and relatively little energy consumption \cite{FRANCESCHINI1992,EXPERT2011,FLOREANO2013,SONG2013}. 

It is well-known that optical flow can convey information on the ratio between distance and velocity, but that additional information is necessary to disentangle these quantities. This `scaling' can be performed by means of maneuvers that have a known range or velocity (as in parallax \cite{KRAL1998}), or with any sensor that directly or indirectly provides distance or velocity estimates, e.g., accelerometers or sonar \cite{LYNEN2013,ENGEL2014}.

Flying insects such as fruit flies do not seem to have scaling sensors \cite{EXPERT2015}. In addition, it is unlikely that they know the range of their maneuvers since they only have access to their air speed, not their ground speed \cite{BREUGEL2014}. Nonetheless, flying insects are able to navigate successfully. They follow straightforward strategies based on optical flow observables. First, it was found that honeybees keep ventral flow (defined as $\vartheta_x = \frac{v_x}{z}$) constant when they perform grazing landings \cite{SRINIVASAN1996}. This strategy was studied for flying robots and simulated spacecraft \cite{RUFFIER2005,ORCHARD2009,VALETTE2010,IZZO2011}. However, these studies required a ``pitching'' law, ultimately depending on an estimate of the height and velocity. The main reason seemed the absence of control of the vertical dynamics. Therefore, more recent landing strategies also include the flow divergence ($D$, typically defined as $\frac{-v_z}{z}$, with the positive $z$, $v_z$ axis pointing up) or time-to-contact ($\tau = \frac{-z}{v_z}$) \cite{herisse2012,IZZO2012B,KENDOUL2014,ALKOWATLY2014}. These strategies either enforce a decreasing time-to-contact (as humans do for car braking \cite{LEE1993}) or keep the divergence constant (as honeybees do for straight landings \cite{BAIRD2013}). Recently, in \cite{EXPERT2015}, ventral flow and divergence were combined with slope estimation and an anemometer in order to fly over uneven terrain, without a need for inertial frames or scaling sensors.  %

The absence of velocity and distance information has been a defining property of the work on optical flow based navigation strategies. However, some observations suggest that for successful navigation distance estimates are still of importance. First, for successful optical flow landing control, the gains of the controller are tuned to a range of specific heights and velocities. For instance, for the constantly decreasing time-to-contact landings in \cite{KENDOUL2014}, the gains are scheduled by means of $\tau$, along a trajectory starting at a specific initial height and velocity. Second, while for a constantly decreasing time-to-contact landing it is possible to perform a final landing procedure (when the time-to-contact is almost zero), it is not obvious how such a procedure should be triggered in a constant time-to-contact or divergence landing. 

It \emph{is} actually possible to estimate distances based on optical flow, if a robot has access to copies of its control input signals (referred to as ``efference copies'' in the biological literature). Two main strategies have been proposed so far in the literature. 

The first strategy for estimating distances is well-established in the field of Image Based Visual Servoing (IBVS). In the task of IBVS a robot has to regulate certain target features to desired image locations, without knowing a priori their dimensions or geometry. Quite early it was observed that if a fixed gain is used to map an image error to a control input, the response of the system depends on the distance to the target \cite{corke1992dynamic}. Given a model of the robot's dynamics, the effects of the robot's actions on the image features can be used to estimate the distance to the target in an adaptive gain control scheme (see, e.g., \cite{asl2014adaptive}). 

The second strategy for estimating distances was recently proposed by \cite{BREUGEL2014}. They showed that during a constant divergence approach of an object, the control input $u$ (the thrust) is proportional to the distance $d$ - and that efference copies can then be used as a stand-in for distance. This solution was tested on a camera mounted on a rail and successful distance estimates were obtained by integrating information during the entire approach.

Both strategies above assume that the robot's accelerations are completely determined by the control inputs. However, flying robots and insects are also subject to significant external accelerations caused by wind or wind gusts. 

The \textbf{main contribution} of this article is the proposition of a novel strategy for distance estimation with optical flow and efference copies. The central idea is to estimate distance by exploiting the self-induced oscillations that result from the fundamental imperfection of fixed-gain optical flow based control. The strategy provides two major advantages with respect to the previously proposed strategies: (1) it works straightforward even in the presence of wind, and (2) it can be used in a broader range of maneuvers, and notably in hover. 

%

The remainder of the article is organized as follows. In Section \ref{section:thrust-based}, the distance estimation method previously proposed in \cite{BREUGEL2014} is explained and investigated in the presence of wind. Subsequently, the new stability-based strategy to distance estimation is proposed in Section \ref{section:stability-based}. In Section \ref{section:distance_estimation} it is shown how self-induced oscillations can be used to trigger a final landing procedure. Subsequently, in Section \ref{section:adaptive_control} adaptive gain control is introduced to determine the distance in hover and to estimate distances along an entire constant divergence landing. In Section \ref{section:robot_experiments}, the findings are empirically verified with real-world experiments using a Parrot AR drone. Finally, the results are discussed in Section \ref{section:discussion} and conclusions are drawn in Section \ref{section:conclusion}. 

%

\section{Thrust-based Distance Estimation} \label{section:thrust-based}
In this section, the distance estimation method proposed in \cite{BREUGEL2014} is explained and investigated in the context of a (robot) landing task. For reasons that will become clear below, this strategy will be referred to as a `thrust-based' method. It is shown that the application of this method is complicated by the presence of wind. 

\begin{figure}[hbt!]
\centering
\includegraphics[width=8cm]{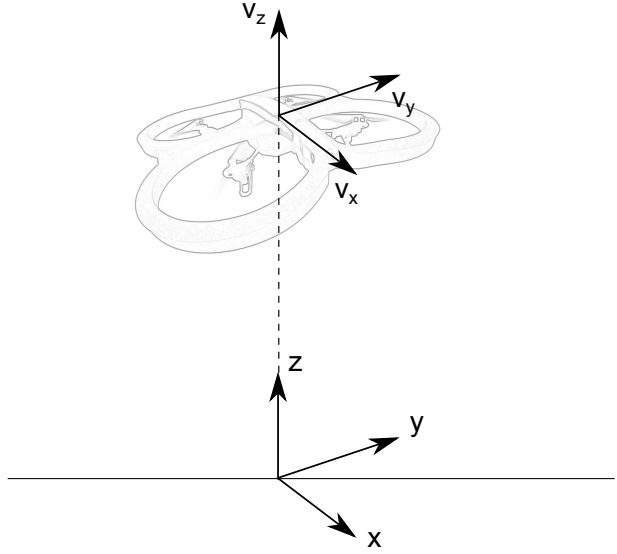}
\caption{Axis system employed in the article.} \label{figure:axes}
\end{figure}

Figure \ref{figure:axes} shows the definition of the axes as used in the formulas. In order to keep the equations as uncluttered as possible, most of the analysis in this article focuses purely on the $z$-axis. Generalizations to other looking and movement directions are easily made (see, e.g., Appendix \ref{section:other_movement_directions}), as are the inclusion of different attitudes or a displacement and rotation of the camera with respect to the body center of mass. 

Importantly, as `visual observable', the variable $\vartheta_z$ is introduced, which is the relative velocity $\frac{v_z}{z}$. It is related to the divergence as $\vartheta_z = -D$, and hence a constant or zero divergence landing is also a constant or zero $\vartheta_z$ landing. The relative velocity $\vartheta_z$ is determined on the basis of the spatial gradient of optical flow in the image, but the algorithms to do this are considered outside the scope of the article and the reader is referred to, e.g., \cite{DECROON2013,ALKOWATLY2014}. 

\subsection{Relation between $u$ and $z$}
As was shown in \cite{BREUGEL2014}, when perfectly following a constant divergence landing, there is a straightforward relation between the control input $u_z$ (thrust) and $z$. Let us start from the visual observable $\vartheta_z$:
\begin{equation} \label{relative_vert_vel}
\vartheta_z = \frac{v_z}{z}
\end{equation} 
If $\vartheta_z$ is differentiated over time, it results in:
\begin{equation}
\dot{\vartheta_z} = \frac{a_z}{z} - \frac{v_z^2}{z^2}
\end{equation} 
\noindent Using Eq. \ref{relative_vert_vel}:
\begin{equation} \label{z_formula}
z = \frac{a_z}{\vartheta_z^2 + \dot{\vartheta_z}}
\end{equation}
So, $z$ is expressed in terms of `observables' ($\vartheta_z$ and its time derivative $\dot{\vartheta_z}$), and the vertical acceleration $a_z$. Adding an accelerometer to the robot will allow for estimation of $z$ (e.g., \cite{IZZO2013}). However, it requires using the time derivative of $\vartheta_z$, which typically induces a lot of noise in the observation.

Van Breugel et al. \cite{BREUGEL2014} describe the equations of motion in a linear state space model as:
\begin{equation} \label{state_space}
\begin{bmatrix} \dot{z}(t) \\ \dot{v_z}(t) \end{bmatrix} = \begin{bmatrix} 0 & 1 \\ 0 & 0 \end{bmatrix} \begin{bmatrix} z(t) \\ v_z(t) \end{bmatrix} + \begin{bmatrix} 0 \\ 1 \end{bmatrix} u_z(t)
\end{equation}
This model entails that $u_z$ = $a_z$. Making this assumption, Eq. \ref{z_formula} becomes:
\begin{equation} \label{z_formula_u}
z = \frac{u_z}{\vartheta_z^2 + \dot{\vartheta_z}}
\end{equation}
In \cite{BREUGEL2014} it has been investigated what would happen if a perfectly constant divergence landing is performed, $\dot{\vartheta_z} = 0$ and $\vartheta_z = -c^2$ ($c^2$ to indicate that $\vartheta_z$ has a negative set point): 
\begin{equation} \label{thrust_based_z}
z = \frac{u_z}{c^4}.
\end{equation}
\noindent Eq. \ref{thrust_based_z} shows that during a perfect constant divergence landing, the thrust $u_z$ is a scaled version of the height. Hence, it can be used as a stand-in for the height. For this reason, the method is referred to in this article as a `thrust-based' method. 

\subsection{Dealing with wind} \label{section:wind}
The assumption $a_z = u_z$ effectively implies that any possible gravity is subsumed under the $u_z$-term:
\begin{equation} \label{a_z}
a_z = -g + \frac{u_z'}{m} = u_z
\end{equation}
\noindent, where $u_z' = m (a_z + g)$ is the actual upward thrust generated by the robot in Newton. Two observations are in place. First, it is common in control system theory to subsume factors such as gravity and the mass into $u_z$, as the calculation of $u_z'$ is straightforward and does not depend on any state variables. Second, it is unlikely that a robot's command signal (and hence efference copy) is equal to $u_z'$. A command signal $u_z''$ (in any unit, e.g., the commanded RPM of a robots' propellors or the flapping frequency of an insect) will typically have to be mapped to $u_z'$ (in Newton) with an actuator effectiveness estimate. If such a mapping is unknown, the estimated height $\hat{z}$ from Eq.~\ref{thrust_based_z} will have a possibly nonlinear relation to $z$: 
\begin{equation}
\hat{z}=g(z).
\end{equation}
\noindent As long as the function $g$ is invertible, this mismatch in unit essentially does not matter. Taking this reasoning into account, for the remainder of the article a copy of $u_z'$ will be referred to as an `efference copy'.  

Eq. \ref{a_z} is only valid if the control force is the only force acting on the robot. In a vacuum environment this can be approximately correct, e.g., for a moon landing. However, a robot flying in the air will undergo additional accelerations. Most important is the drag force, which depends on the robot's movement relative to the air surrounding it:
\begin{equation} \label{drag_force}
f_D = \textrm{sign}(v_{\textrm{air}}) \frac{1}{2} \rho C_D A v_{\textrm{air}}^2,
\end{equation}  
\noindent where:
\begin{equation}
v_{\textrm{air}} = v_{\textrm{wind}} - v_z,
\end{equation}
\noindent and $\textrm{sign}(v_{\textrm{air}})$ indicates the directionality of $f_D$ along the $z$-axis. This leads to an additional acceleration:
\begin{equation} \label{az_drag}
a_z = -g + \frac{u_z' + f_D}{m} 
\end{equation}
\noindent, where $f_D$ is a time-varying value involving a non-linear function of $v_z$ and an uncontrolled variable $v_{\textrm{wind}}$. 

\begin{figure}[htb]
\centering
\includegraphics[width=10cm]{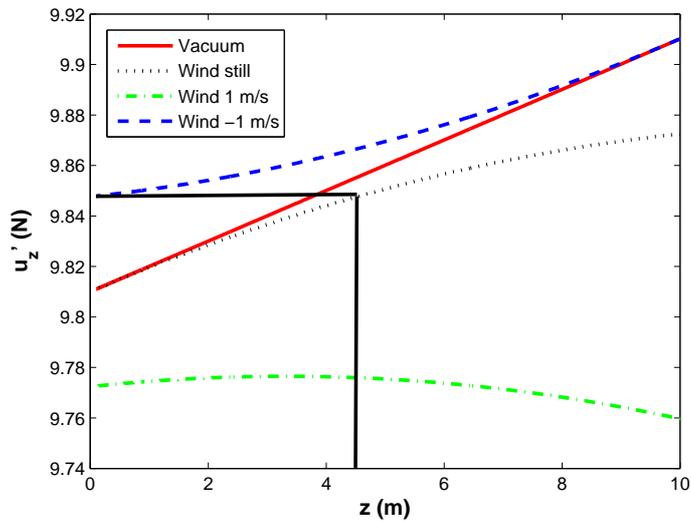}
\caption{Thrust $u_z'$ versus height $z$ in different environmental conditions (see legend). Two black lines illustrate an estimation error when the wind is unknown. The lines indicate that the thrust with a downward wind of $-1$m/s at $z=0$m corresponds to a thrust in a wind-still environment at a height of $z\approx4.5$m .}\label{figure:perfect_landings}
\end{figure}

It is informative to study the effect of drag and wind on the required $u_z'$, when the robot follows a \emph{perfect} constant divergence landing. In such a landing, $\vartheta_z = -c^2$ at every point by definition. The acceleration resulting from drag and the required $u_z'$ are calculated to match. Figure \ref{figure:perfect_landings} shows the value of $u_z'$ for a robot with assumed values of $m=1$kg, $\rho=1.204$, and $g=9.81$m/s$^2$. Furthermore, $C_D=0.25$, and $A=0.25$, which are rather conservative values representing for instance a small quad rotor. The figure shows the thrust $u_z'$ versus height $z$ in different environmental conditions: vacuum (red solid line), wind still (black dotted), in a downward wind of $-1$m/s (blue dashed), and in an upward wind of $1$m/s (green, dashed-dotted). 

Three main observations can be made from Figure \ref{figure:perfect_landings}: (1) the red solid line in vacuum indeed indicates the derived linear relationship between $u_z$ and $z$, (2) in a wind-still environment, this relationship is nonlinear, but invertible - and hence as useful, (3) adding a modest wind speed to the equation already makes significant differences to $u_z'$. Two black lines indicate that the thrust with a downward wind of $-1$ at $z=0$m corresponds to a thrust in a wind-still environment at a height of $z\approx4.5$m. The curve for a wind of $1$m/s does not even match that in a wind-still environment. Hence, even with a perfect constant divergence landing these differences distort the relationship between $u_z$ and $z$. 

So, in order to retrieve the right $u_z$-curve (see Figure \ref{figure:perfect_landings}), the wind will have to be measured. Flying insects may be able to measure the wind velocity with their hairs \cite{fuller2014flying}, while robots can be equipped with an anemometer, as in \cite{EXPERT2015}. Although measuring the wind velocity is a possibility, this does not lead to an easy solution of the distance estimation problem. First and foremost, controlling a constant divergence landing in a real system means dealing with sensing and actuation delays, visual inaccuracies, and changing external factors such as wind gusts. Any real system will thus deviate from the perfect landing profile, leading to the necessity of command signal variation and hence a varying $u_z'$. Indeed, this effect can be seen in the raw $\hat{z}$ estimates in \cite{BREUGEL2014} for the camera-system mounted on a rails. In \cite{BREUGEL2014} they tackled this issue with a robust estimation scheme that integrates information during the entire approach. The problems are much worse though for a freely flying system, which is also subject to wind.

\section{Instability of Constant Divergence Landings} \label{section:stability-based}
The onset of the work in this article came from the difficulty of making optical flow based landing strategies work on real Micro Air Vehicles (MAVs). The study in \cite{KENDOUL2014} is especially instructive, since multiple control laws are studied, of which the gains depend to different extents on the initial conditions. Especially important are the remarks and observations on the difficulty of getting the optical flow control to work close to the landing surface, even suggesting to disengage the optical flow control under a threshold distance (as is also done in, e.g., \cite{VALETTE2010}). 

Please note that it is not easy to identify the cause of oscillations close to the surface when performing vision-based control with a real MAV. Oscillations close to the ground can potentially be explained by the ``ground effect'' due to the MAV's downwash and by the vision process (more blurry images, larger optical flow vectors that are harder to track). Below it will become clear though that even without such additional effects, the instability problem arises.

Based on the dependence of the gains on the actual height and velocity, in this article a novel strategy is proposed for distance estimation: \emph{stability-based} distance estimation. In this section, first the fundamental reason for the gain tuning problem of optical flow based control is explained (Subsection \ref{section:fundamental_reason}). Subsequently, a specific control law is assumed and it is shown analytically that the stability directly relates the control gain to the height (Subsection \ref{section:control_stability}).

\subsection{Fundamental reason for gain tuning} \label{section:fundamental_reason}
In order to see why a single gain is only stable for a given range of heights, let us go back to Eq. \ref{z_formula}, and reorganize it as:
\begin{equation} \label{z_formula2}
\dot{\vartheta_z} = \frac{a_z}{z} - \vartheta_z^2 
\end{equation}
\noindent Assuming $a_z$ as in Eq. \ref{az_drag}:
\begin{equation}
\dot{\vartheta_z} = \frac{-g + \frac{u_z' + f_D}{m}}{z} - \vartheta_z^2 
\end{equation}
\noindent Differentiating this formula with respect to $u_z'$ gives:
\begin{equation} \label{fundament}
\frac{\partial \dot{\vartheta_z}}{\partial u_z'} = \frac{1}{mz}
\end{equation}

Eq. \ref{fundament} shows that a change in thrust has a larger effect on the change in $\dot{\vartheta_z}$ close to the landing surface than far away (independently of external accelerations). The constancy of this relation depends on the constancy of the robot's mass, which applies for landing quad rotors but not for spacecraft. A spacecraft will loose more and more mass when landing, hence aggravating the effect in Eq. \ref{fundament}.

Thus, a control gain that leads to a satisfactory control performance far away from the landing surface, will lead to very large effects on $\dot{\vartheta_z}$ and hence indirectly on $\vartheta_z$ close to the landing surface. Although theoretically one could use $\frac{\partial \dot{\vartheta_z}}{\partial u_z'}$ directly for distance estimation ($z = \frac{\partial u_z'}{\partial \dot{\vartheta_z}} \frac{1}{m}$), in practice this value is extremely noisy, since it involves a partial derivative of the already noisy value $\dot{\vartheta_z}$. 

\subsection{Relation between control stability, gain, and height} \label{section:control_stability}
In this subsection, it will be shown that the larger influence of $u_z$ on $\vartheta_z$ at lower heights eventually leads to instability at a specific height. For the analysis, the following constant divergence control law is studied, as was used in \cite{BREUGEL2014}:
\begin{equation}
u_z = K_z (\vartheta_z^{*} - \vartheta),
\end{equation}
\noindent where it is easy to see that in a noiseless, delayless system:
\begin{equation} \label{continuous_stable}
\lim_{K_z \to \infty} \frac{u_z}{K_z} = (\vartheta_z^{*} - \vartheta_z) = 0,
\end{equation}
\noindent and hence $\vartheta_z = \vartheta_z^{*}$. Real systems always have some noise and delay. Below, it will be shown that just discretizing the control with Zero-Order-Hold (ZOH) will already introduce an instability into the system.

Let us assume the state space model of Eq. \ref{state_space}, ignoring drag for the moment. The corresponding observation is:
\begin{equation}
y = \vartheta_z = \frac{v_z}{z},
\end{equation}
\noindent which is a nonlinear function. Linearizing the state space model gives:
\begin{equation}
\Delta y(t) = \begin{bmatrix} \frac{-v_z}{z^2} &  \frac{1}{z} \end{bmatrix} \begin{bmatrix} \Delta z(t) & \Delta v_z(t) \end{bmatrix}^{\top},
\end{equation}
\noindent so that the state space model matrices are:
\begin{equation} \label{ss_continuous}
A = \begin{bmatrix} 0 & 1\\ 0 & 0 \end{bmatrix} \: \: B = \begin{bmatrix} 0 \\ 1 \end{bmatrix} \\
C = \begin{bmatrix} \frac{-v_z}{z^2} &  \frac{1}{z} \end{bmatrix} \: \: D = [0].
\end{equation}

As mentioned, the continuous system without noise and delay is stable (Eq. \ref{continuous_stable}). Here, the discretized system is studied, which has the following state space model matrices corresponding to the continuous ones in Eq. \ref{ss_continuous}:
\begin{equation} \label{ss_discrete}
\Phi = \begin{bmatrix} 1 & T\\ 0 & 1 \end{bmatrix} \: \: \Gamma = \begin{bmatrix} \frac{T^2}{2} \\ T \end{bmatrix} \\
C = \begin{bmatrix} \frac{-v_z}{z^2} &  \frac{1}{z} \end{bmatrix} \: \: D = [0],
\end{equation}
\noindent where $T$ is the discrete time step in seconds. The transfer function of the open loop system can be determined to be:
\begin{equation}
G(w) = C (w I - \Phi)^{-1} \Gamma,
\end{equation}
\begin{equation}
= \frac{(zT - \frac{1}{2}v_zT^2)w - zT - \frac{1}{2}v_zT^2}{z^2(w-1)^2},
\end{equation}
\noindent where we use $w$ as the $Z$-transform variable, since $z$ already represents height. The feedback transfer function is:
\begin{equation} \label{general_transfer_function}
G(w) = \frac{K_z \left( (zT - \frac{1}{2}v_zT^2)w - zT - \frac{1}{2}v_zT^2 \right)}{z^2(w-1)^2 + K_z \left( (zT - \frac{1}{2}v_zT^2)w - zT - \frac{1}{2}v_zT^2 \right)}.
\end{equation}
Eq. \ref{general_transfer_function} shows that, given a gain $K_z$ and time step $T$, the dynamics and stability of the system depend both on the height $z$ and velocity $v_z$. 

\begin{figure}
\centering
\includegraphics[width=8cm]{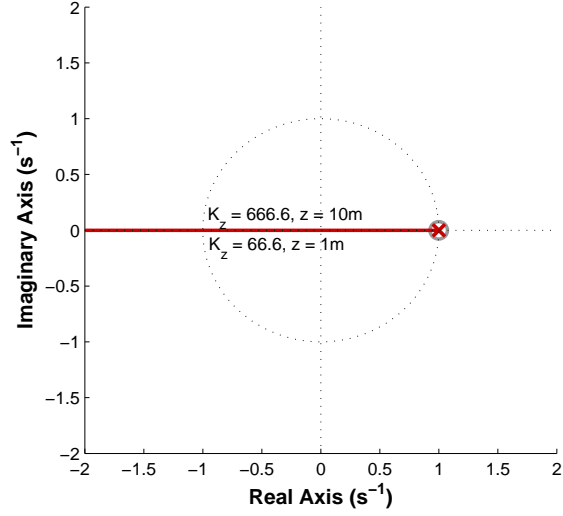}
\caption{Root locus plots of a ZOH-model with $T=0.03$s for $z=1$m and $z=10$m. The plots are equal in shape, but for each point $K_z$ is different. This is indicated for the (approximations to) the values of $K_z$ at which the system becomes unstable ($w=-1$).} \label{figure:rlocus_dv}
\end{figure}

Figure \ref{figure:rlocus_dv} is the root locus plot of $G(w)$. For the $Z$-transform any pole in the unit circle is a stable pole. The different elements of the plot can be found back in Eq. \ref{general_transfer_function}. The numerator indicates a single finite zero at:
\begin{equation}
w_0 = \frac{zT + \frac{1}{2}v_z T^2}{zT - \frac{1}{2}v_z T^2}.
\end{equation}
\noindent Given $z>0$, $T>0$, and $v_z<0$, $w_0$ is positive and due to the typically small $T$ values slightly smaller than 1. There is also a negative infinite zero. Since the denominator of  Eq. \ref{general_transfer_function} is a second order equation in $w$, $G(w)$ has two poles, both of which are stable. For $K_z=0$, the poles are located at $w=1$. As the gain $K_z$ increases, one pole moves toward the finite zero and one toward the infinitely negative zero. This latter pole is interesting, because as soon as it passes $w=-1$, the system becomes unstable. Setting $w=-1$ in the denominator and equating it with $0$ gives an equation for the unstable gain value $K_z$:
\begin{equation} \label{vacuum_instable_K}
K_z = \frac{2}{T}z.
\end{equation}

This implies that given a specific gain $K_z$ and time step $T$, \emph{there is always a height at which the control system will become unstable}:
\begin{equation} \label{vacuum_z_unstable_K}
z = \frac{1}{2} K_z T.
\end{equation}
When using a fixed gain, a tradeoff has to be made between the control performance at a larger height (requiring a large gain) and at lower heights (requiring a small gain). Please remark that the instability at lower altitudes is exactly what is observed for robotic systems performing constant divergence landings! 

\subsection{Including wind}
In this subsection it is shown that also for a model with wind, the instability of the system given a $K_z$ depends on $z$. Including wind leads to a changed formula for $\dot{v_z}$ (see Eqs. \ref{drag_force} and \ref{az_drag}): 
\begin{equation}
\dot{v_z} = u_z + \textrm{sign}(v_{\textrm{wind}} - v_z) \frac{1}{2m} \rho C_D A (v_{\textrm{wind}} - v_z)^2
\end{equation}
\noindent, where from now on $\beta$ will stand in for the constant $\frac{1}{m} \rho C_D A $. In order to obtain a linear state space model, this equation is linearized to obtain:
\begin{equation}
\Delta \dot{v_z} = \Delta u_z - \left( \textrm{sign}(v_{\textrm{wind}} - v_z) \beta (v_{\textrm{wind}} - v_z) \right) \Delta v_z
\end{equation}
\noindent, where the factor multiplied with $\Delta v_z$ is a constant in the linearized model ($v_{\textrm{wind}}$ and $v_z$ are given the values at the linearization point). In order to avoid cluttering the formulas, this constant is represented by $p = \textrm{sign}(v_{\textrm{wind}} - v_z) \beta (v_{\textrm{wind}} - v_z)$ (where $p > 0$). This leads to the following continuous, linear state space model:
\begin{equation} \label{ss_wind_continuous}
A = \begin{bmatrix} 0 & 1\\ 0 & -p \end{bmatrix} \: \: B = \begin{bmatrix} 0 \\ 1 \end{bmatrix} \\
C = \begin{bmatrix} \frac{-v_z}{z^2} &  \frac{1}{z} \end{bmatrix} \: \: D = [0],
\end{equation}
\noindent which results in the $Z$-transform matrices:
\begin{equation}\label{ss_wind_discrete}
\Phi = \begin{bmatrix} 1 & \frac{(1-e^{-p T})}{p}\\ 0 & e^{-p T} \end{bmatrix} \: \: \Gamma = \begin{bmatrix} \frac{T}{p} - \frac{(1-e^{-p T})}{p^2} \\ \frac{(1-e^{-p T})}{p} \end{bmatrix}
\end{equation}
\begin{equation}
C = \begin{bmatrix} \frac{-v_z}{z^2} &  \frac{1}{z} \end{bmatrix} \: \: D = [0],
\end{equation}
\noindent and a rather involved transfer function $H(w)$. The root locus plot of $H(w)$ is very similar to that of $G(w)$, also with a negative infinite zero. Following the same procedure as for the `vacuum' model, setting $w=-1$ and equating the denominator with 0, gives:
\begin{equation} \label{drag_instable_K}
K_z = \frac{(2p^2 + 2 p^2e^{p T})z^2}{(2e^{pT}-2-Tp-Tpe^{pT}) v_z + (2pe^{pT} - 2p) z}.
 \end{equation} 
\noindent Eq. \ref{drag_instable_K} includes many instances of $p$ that depends on $v_z$ and $v_{\textrm{wind}}$, and also contains a term in the denominator with $v_z$. This suggests that there is no fixed linear relation between $K_z$ and $z$. However, closer inspection shows that the term $(2e^{pT}-2-Tp-Tpe^{pT}) \approx 0$. Rearranging terms, this leads to:
\begin{equation} \label{drag_instable_K}
K_z = \frac{2p^2 + 2 p^2e^{p T}}{2pe^{pT} - 2p} z,
\end{equation} 
\noindent which still depends on $p$. It turns out that the fraction that is multiplied with $z$ is almost identical to the fraction $\frac{2}{T}$. Indeed, solving Eq. \ref{drag_instable_K} for different variable settings gives almost identical results to Eq. \ref{vacuum_instable_K}.

\subsection{Continuous system with drag and delay} \label{section:continuous_delay}
\begin{figure}
\centering
\includegraphics[width=8cm]{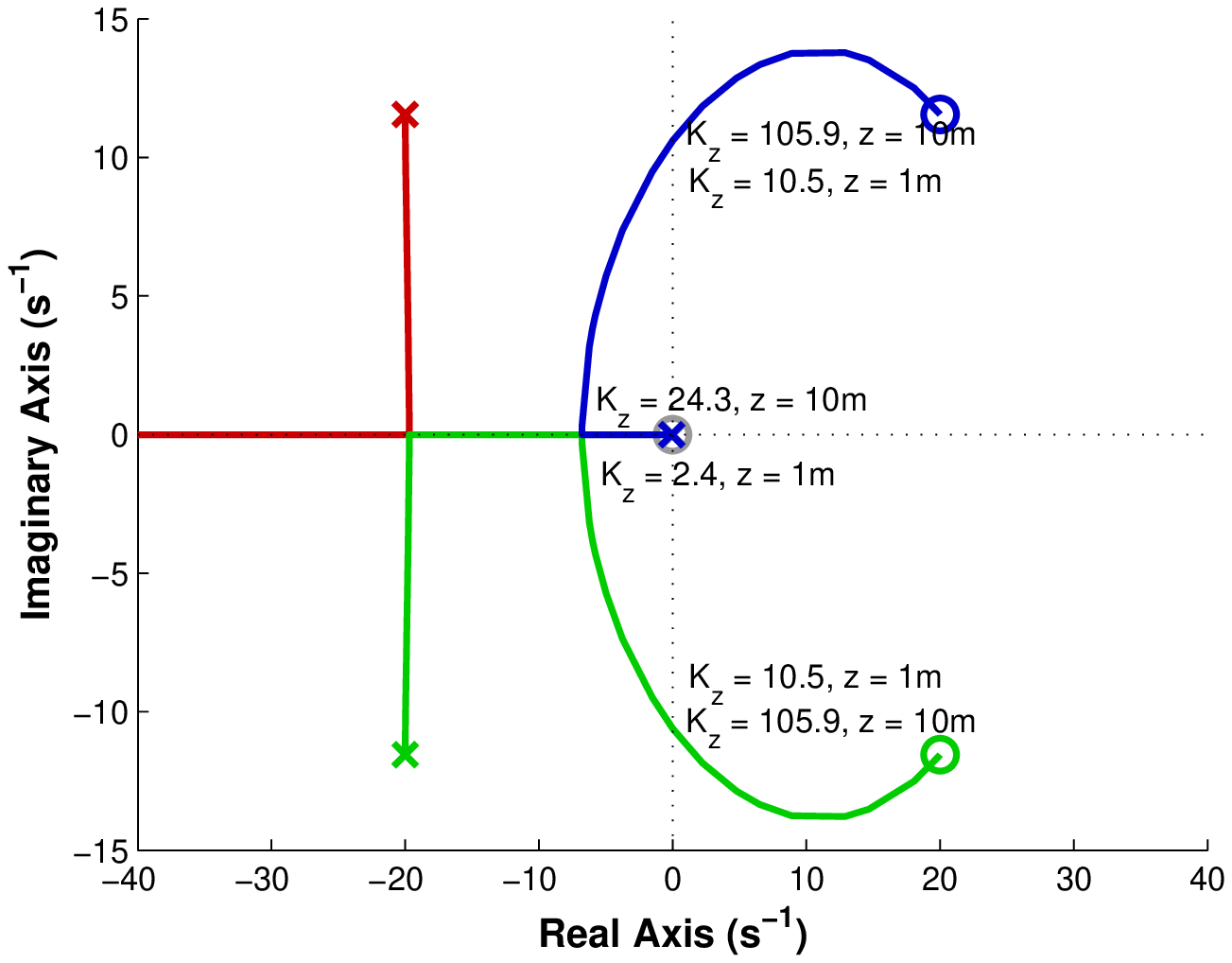}
\caption{Root locus plots of a continuous model with a delay of $T=0.15$s (as approximated by a second order Pad\'{e} transfer function), and aerodynamic drag with $\frac{1}{2} \rho C_D A = 0.5$, for $z=1$m and $z=10$m. The plots at different heights are equal in shape, but for each point in the plot $K_z$ is different. This is indicated for the (approximations to) the values of $K_z$ at which the poles get an imaginary component (start of oscillations), and where they cross the imaginary axis (become unstable).} \label{figure:cont_rlocus}
\end{figure}

The previous subsections discussed discretized models for which unstable $K_z$ exist, and where the limit case can be analytically expressed as a function of $z$. In order to show that continuous systems encounter the same type of phenomenon, Figure \ref{figure:cont_rlocus} shows the root locus plot of a continuous system with wind, a delay of $\Delta t = 0.03$s and $c^2 = 0.01$. The delay introduces two zeros in the (unstable) right-hand plane of this continuous root locus plot. A few values of $K_z$ are plotted, where the poles get an imaginary component and where they cross the imaginary axis (shown for $z=10$m and $z=1$m). 

\section{Stability-based Distance Estimation} \label{section:distance_estimation}
In the previous section, it was established that the instability of a constant divergence landing system depends on the height. If this instability can be detected by a robot or insect, it can be used for distance estimation!

\subsection{Detection of Self-Induced Oscillations}  \label{section:detection}
The instability discussed in Section \ref{section:control_stability} is induced by the robot itself. Before the system gets unstable, it will start to show oscillations. This can be seen for instance in Figure \ref{figure:cont_rlocus}. When $K_z= 24.3$, the control system's poles will start to have an imaginary component and hence oscillate around the desired value from $z=10$m downward. In the field of aerospace, such oscillations are a well-known problem, and referred to as Pilot-Induced Oscillations (PIO). 

There have been several investigations on the automatic, on board detection of self-induced oscillations \cite{miao1999automatic,Dimogianopoulos2005,rzucidlo2007detection,yildiz2011control,chowdhary2011frequency}. The two most important properties of self-induced oscillations are that (1) there is a phase shift between the observations and control inputs in the order of $90$ - $180^{\circ}$, and (2) the oscillations are of a significant magnitude. A typical detection method involves a Fast Fourier Transform (FFT) of the observations and control inputs in order to detect the occurrence of these properties \cite{rzucidlo2007detection,chowdhary2011frequency}. In this article, in the interest of a computationally efficient and straightforward method, we will investigate the use of the covariance between the control input $u_z'$ and the relative velocity $\vartheta_z$ and the use of $\vartheta_z$ itself. 

\begin{figure}[t]
\centering
\includegraphics[width=7cm]{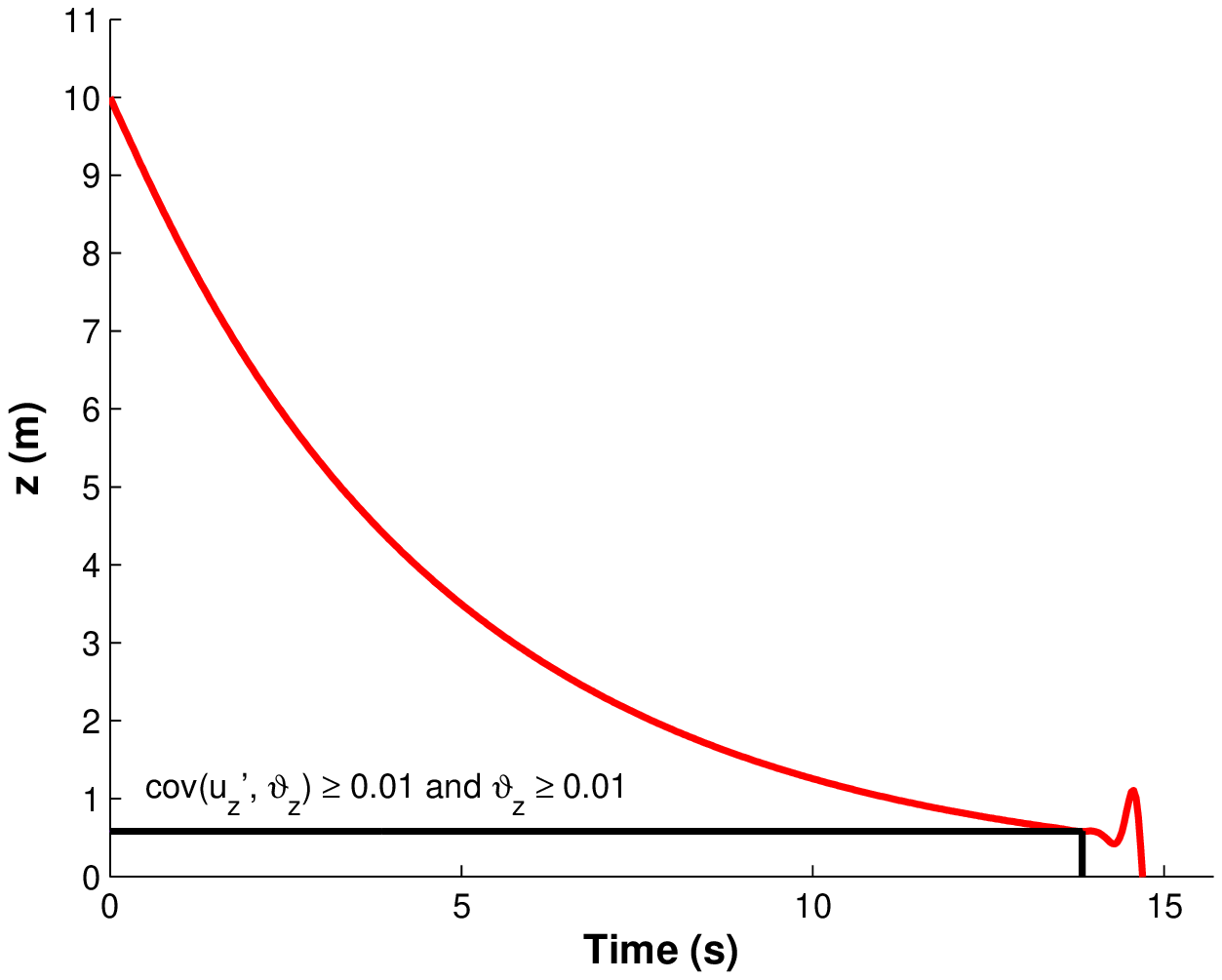} \includegraphics[width=7cm]{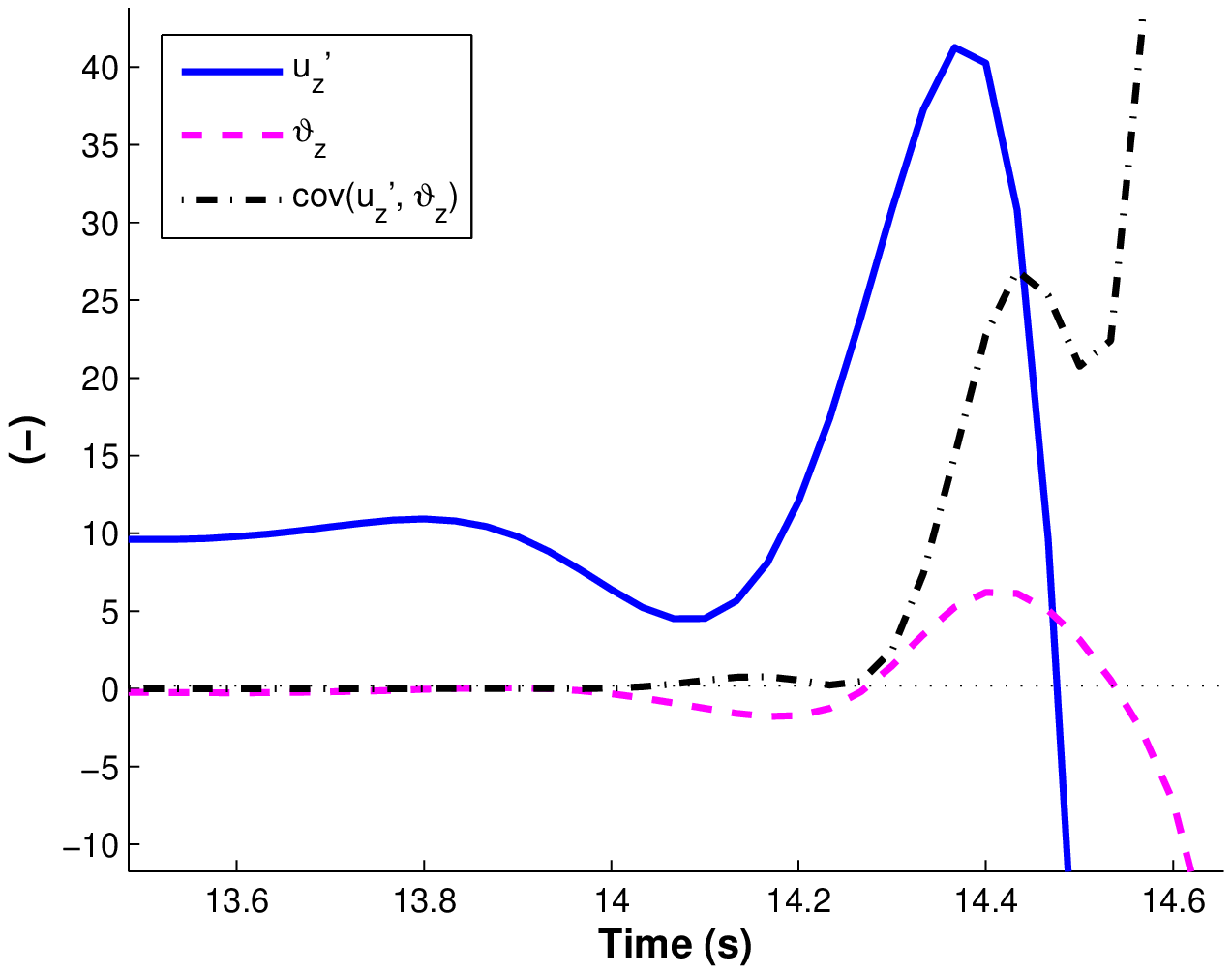}
\caption{\textbf{Left:} Landing in simulation with $K_z=20, c^2=0.2$. The red line is the height, the black line is the point at which the onset of oscillations is detected with the robot's observables. \textbf{Right:}. Three of the robot's observables: $\vartheta_z$ (magenta, dashed), the thrust $u_z'$ (blue, solid), and their covariance $\mathrm{cov}(u_z', \vartheta_z)$ (black, dashed-dotted).} \label{figure:detection}
\end{figure}

As a first test, landings in simulation will be performed. For the simulations, a ZOH-model is assumed with a time step $T = 0.03$s, $\frac{1}{2} \rho C_D A = 0.5$, a mass of 1 kg, and a delay of $\Delta t = 0.15$s (as in Section \ref{section:continuous_delay})\footnote{Please note that while the theoretical analysis involved a linearization of the observation function, the simulation employs the nonlinear observation function (and hence can validate some of the conclusions from Section \ref{section:stability-based}).}. The left part of Figure \ref{figure:detection} shows a landing for $K_z = 20$, $z_0 = 10$, $v_{z0}=-2$, $c^2=0.2$, no wind. It shows that the landing goes smoothly, until the very end at $\approx 0.7$m at which point the robot's height starts to oscillate. 

The right part of Figure \ref{figure:detection} shows values that can be observed by the robot itself, zooming in at the few seconds before landing. The magenta dashed line is $\vartheta_z$. Toward the end of the landing $\vartheta_z$ starts deviating significantly from its reference point $-0.2$ (indicated with the dotted black line). It even obtains positive values, which means that the robot goes up. This itself could be used for the detection of instability, were it not that a wind gust can also move the robot upward, causing a positive $\vartheta_z$ at a larger height. The figure also shows the thrust $u_z'$ (blue solid line) and the covariance of $u_z'$ and $\vartheta_z$ over the previous 20 time steps (black dashed-dotted line). During the smooth part of the trajectory $\mathrm{cov}(u_z', \vartheta_z)$ is a small negative number: when the robot descends too fast ($\vartheta_z$ too negative) the robot thrusts more up, and vice versa. 

Towards the end of the landing the observable $\vartheta_z$ starts to change too quickly for the control system and $\mathrm{cov}(u_z', \vartheta_z)$ becomes positive: if the robot descends too fast ($\vartheta_z$ too negative) the robot will thrust \emph{less} up, and thus it starts inducing oscillations.
Both $\vartheta_z$ and $\mathrm{cov}(u_z', \vartheta_z)$ can be used for detecting the onset of oscillations. The black line in the left part of Figure \ref{figure:detection} shows the first point during the landing at which both variables are higher than $0.01$. 

\begin{figure}[t]
\centering
\includegraphics[width=7cm]{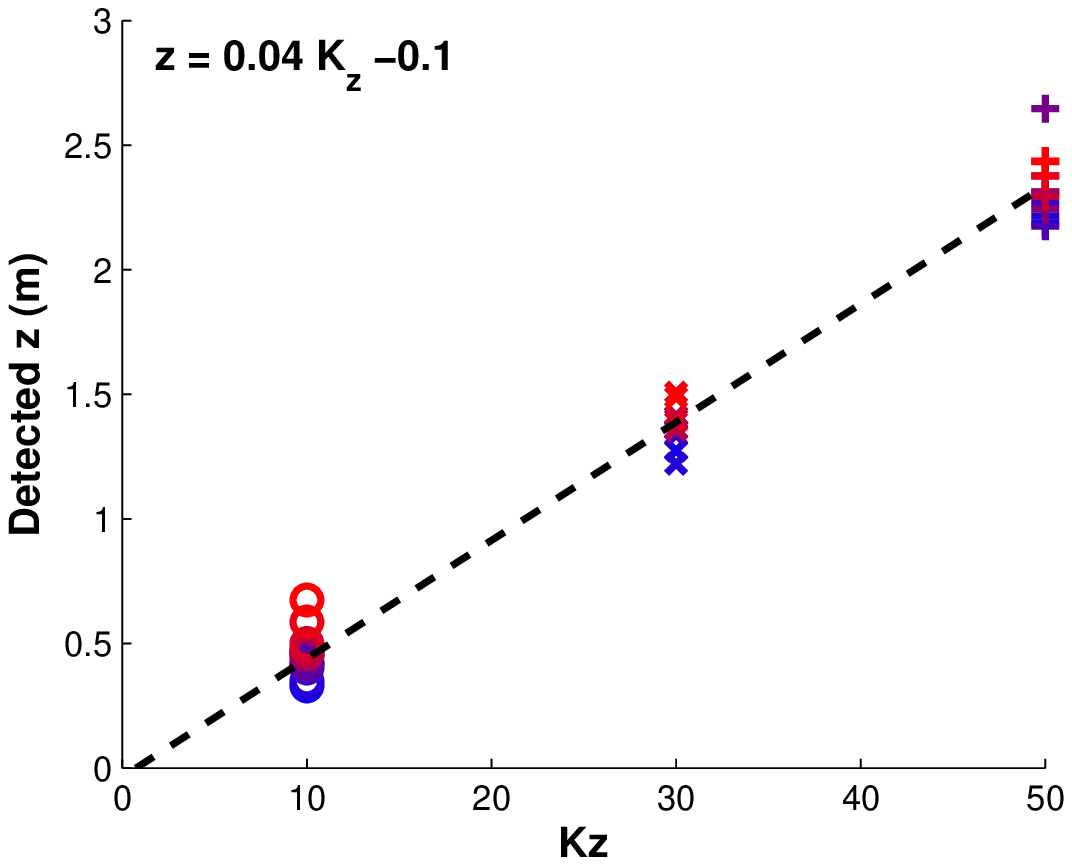} \includegraphics[width=7cm]{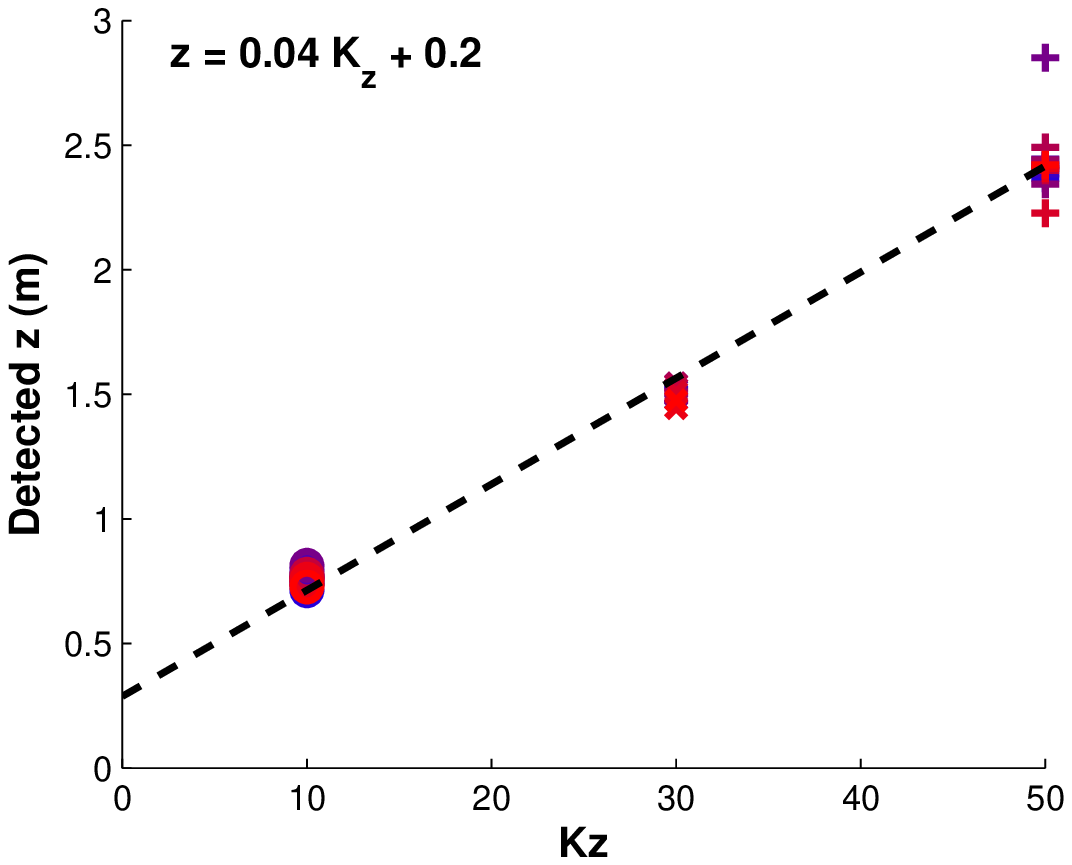}
\caption{\textbf{Left:} Oscillation detection heights for different gains $K_z$ for different wind velocities. The color of the marker represents the wind velocity, ranging from $-3$m/s (blue) to $3$m/s (red). The black dashed line is a linear fit through the detection heights. \textbf{Right:} Results for experiments with a $PI$-controller, with $I_z=1$ as the $I$-gain.} \label{figure:detected_heights}
\end{figure}

In order to test the hypothesis that the onset of oscillations is related to the magnitude of the gain and the height of the robot, simulation runs were performed for gains $K_z \in \{10, 30, 50\}$ and for various types of wind $v_{\mathrm{wind}} \in \{-3, -2.5, \ldots, 2.5, 3\}$ m/s. Furthermore, $z_0 = 20$, $v_{z0}=-1$, $c^2=0.05$. The left part of Figure \ref{figure:detected_heights} shows the results of this experiment, with the different $K_z$ on the $x$-axis and the corresponding $z$ at which self-induced oscillations are detected on the $y$-axis. Each $K_z$ additionally is represented with a different marker, and the markers are colored according to the wind velocity, from blue ($-3$m/s) to red ($3$m/s). The black dashed line is a linear least-squares fit through the points, with as parameters: $z = 0.04 K_z - 0.1$.

There are three main observations. First, as the linear fit shows, higher gains result in oscillations further away from the landing surface. Hence, a fixed gain will result in self-induced oscillations around a certain height. The detection of these oscillations can be used for instance for triggering landing responses (such as the leg extension by fruitflies). Second, a considerable wind difference between $-3$ to $3$ m/s leads to a - for MAVs - relatively small difference of $\approx 0.50$m in the height at which self-induced oscillations occur. Third, the colors show that there is a slight correlation between the detection height and the wind velocity. Analyzing the landings with wind shows that for the higher wind velocities, the control system has more difficulties tracking $\vartheta_z^{*}$, leading to a steady-state error. This suggests that introducing an integrator term could resolve this problem. The results of a $PI$-controller, with $K_z$ as the $P$-gain and $I_z=1$ as the $I$-gain, are shown in the right part of Figure \ref{figure:detection}. Since the $PI$-controller can cancel the steady-state error, the reference $\vartheta_z^{*}$ is tracked much better. As expected, self-induced oscillations still occur, and the detection heights are much closer together. Moreover, inspection of the marker colors shows that there is no obvious relation anymore between the wind velocity and the detection height. In order to keep in line with the system that was analyzed theoretically, for the remainder of the article we will focus on the $P$-controller setup with gain $K_z$.

\subsection{Wind Gusts and Actuator Efficiency} \label{section:wind_gusts}
In the previous subsection, the feasibility of detecting self-induced oscillations for distance estimation was shown. However, two factors were not modeled that could have a significant influence on the practicality of the approach. 

First, during a typical landing outdoors the wind is not constant. The wind can vary and sudden wind gusts can occur. In order to study a rather extreme case, wind gusts are added to the simulation in the form of a sine function:
\begin{equation}
v_{\textrm{gust}} = W \textrm{sin}(a t),
\end{equation}
\noindent where $W$ determines the magnitude of the gusts and $a$ the period. 

Second, as mentioned in Subsection \ref{section:wind}, the mapping from command signal $u_z''$ to the actual force in Newton $u_z'$ depends on the actuator effectiveness. In flying robots such as rotorcraft or flapping wing vehicles this actuator effectiveness, in turn, depends on the air flow. Here, a rotorcraft is assumed and the relation between actuator effectiveness and air flow is modeled according to findings on propellors in \cite{theys2014wind}. Specifically, the following formula was used:
\begin{equation}
u_z' \leftarrow \textrm{max}\{ u_z' - b v_{\textrm{air}} u_z' - c v_{\textrm{air}}, 0 \},
\end{equation}
\noindent so that both the offset and slope of the efficiency change over different air velocities.

\begin{figure}[t]
\centering
\includegraphics[width=7cm]{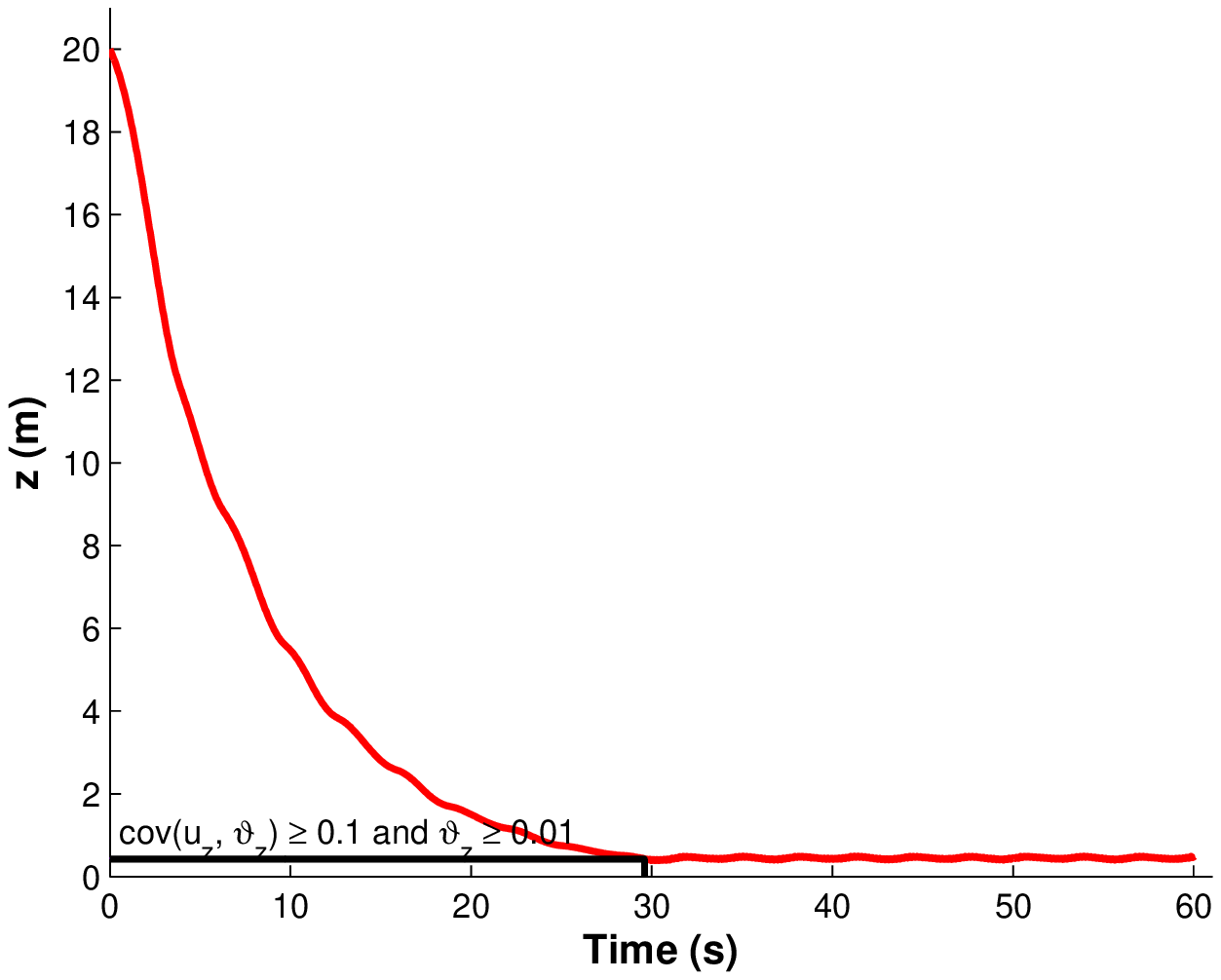} \includegraphics[width=7cm]{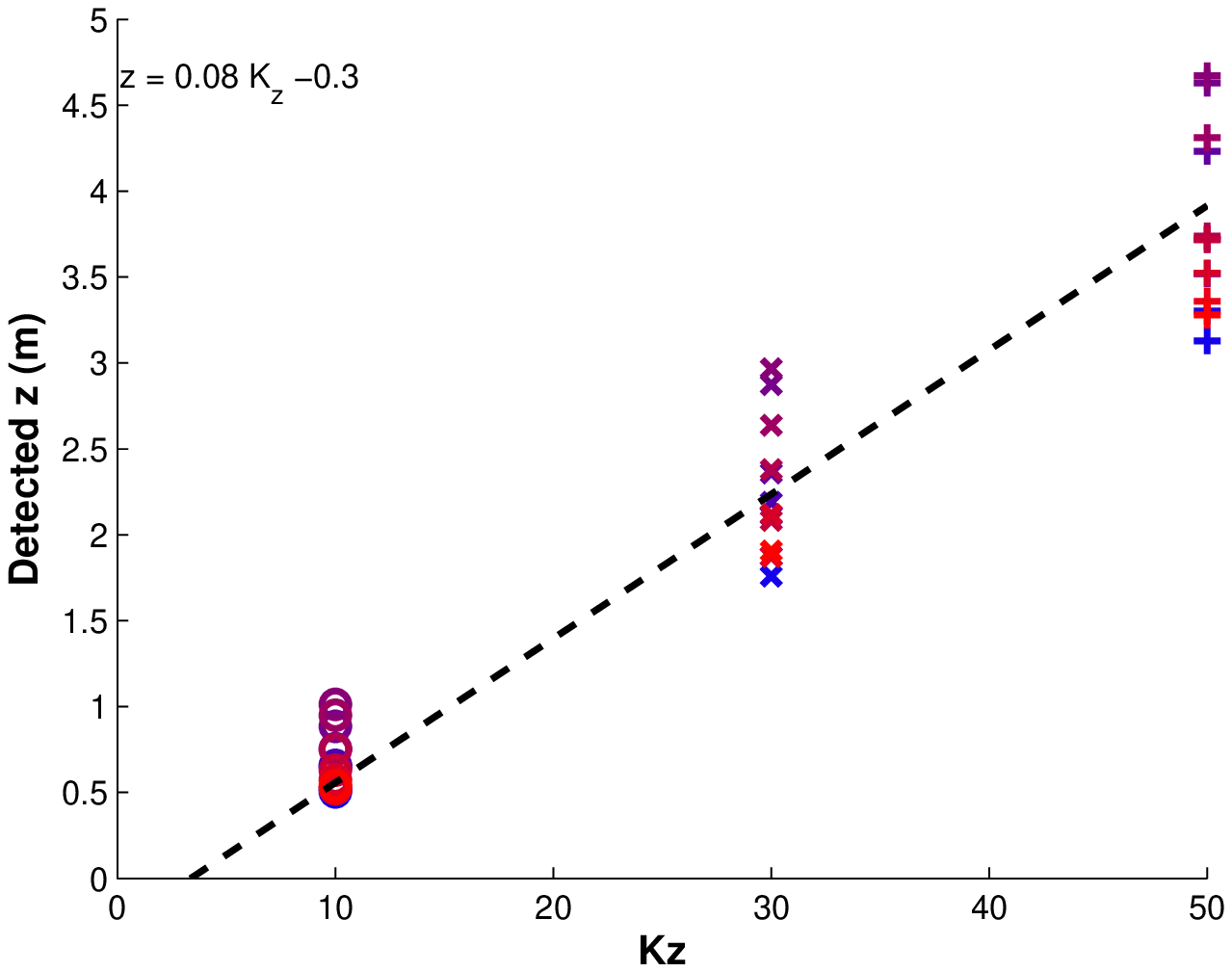} 
\caption{\textbf{Left:} Landing in gusty wind conditions (ranging from $-4$ to $4$m/s in $\pi$ seconds. The red line is the height over time, the black line the detection height. \textbf{Right:} Oscillation detection heights for different gains $K_z$ for different wind velocities in gusty wind conditions. The color of the marker represents the wind velocity, ranging from $-3$m/s (blue) to $3$m/s (red). The black dashed line is a linear fit through the detection heights.} \label{figure:gusts}
\end{figure}

The left part of Figure \ref{figure:gusts} shows a landing with a rather extreme scenario, in which the average wind velocity is $0$m/s, $W=4$, ,$a = 1$, $b=0.5$, and $c=0.5$. This means that per every $\pi$ seconds, the wind varies from $-4$ to $4$m/s! For this gusty scenario, the threshold of $\textrm{cov}(u_z', \vartheta_z)$ is set slightly higher, to $0.1$. The red line in the figure is the height over time, the black line indicates the height at which self-induced oscillations are detected. The right part shows the results over many landings with the same settings but different wind velocities, from $-3$ (blue) to $3$m/s (red) - to which the gusts are added during each landing. The black dashed line is a linear fit through all the detection heights and gains $K_z$. 

The results show that despite the very gusty conditions and varying actuator efficiency, there is still a positive linear relation between $z$ and $K_z$. The uncertainty is slightly higher though than for a constant wind and actuator efficiency, especially for larger gains / heights. 



\section{Adaptive Gain Control} \label{section:adaptive_control}

\subsection{Distance Estimation in Hover} \label{section:hover}
In principle, the detection of self-induced oscillations could be sufficient for flying robots (or insects) to determine when they are close to their landing target. This allows for the triggering of a final landing procedure and hence is behaviorally very relevant. However, it would be extremely useful if the robot could determine height at any distance to the landing surface. 

Surprisingly, the proposed stability-based strategy for height estimation does not require the robot to land! Eq. \ref{vacuum_z_unstable_K} (derived for the vacuum condition) does not depend on $v_z$, nor on $c^2$. This suggests a strategy for determining the height around hover: the robot can set $\vartheta^{*}=0$ and change its gain $K_z$ so that the control loop starts exhibiting small self-induced oscillations. The gain at such a point is then a stand-in for the height.

Specifically, a control law can regulate $\mathrm{cov}(u_z', \vartheta_z)$ by continuously adapting the gain $K_z$. This leads to the following adaptive gain control setup. An inner loop uses $u_z = K_z (\vartheta_z^{*} - \vartheta_z)$, while an additional outer loop controls $K_z$:
\begin{equation} \label{PI-control-Kz}
K_z(t) = K_z'(t) + (P \: K_z'(t)) e_{\mathrm{cov}}(t),
\end{equation}
\begin{equation} \label{PI-control-Kz-prime}
K_z'(t) \leftarrow K_z'(t-T) + (I \: K_z') e_{\mathrm{cov}}(t-T),
\end{equation}
\begin{equation} \label{adaptive_error}
e_{\mathrm{cov}}(t) = \mathrm{cov}(u_z', \vartheta_z)^{*} - \mathrm{cov}(u_z', \vartheta_z)(t),
\end{equation}
\noindent where $P,I \in [0,1]$ are a proportional and integral gain for the outer loop control, both relative to the current $K_z'$. The reason for this is that larger heights involve higher speeds, so that $K_z$ should be changed more quickly for larger $K_z'$ than for smaller $K_z'$. 

\begin{figure}%
\centering
\includegraphics[width=8cm]{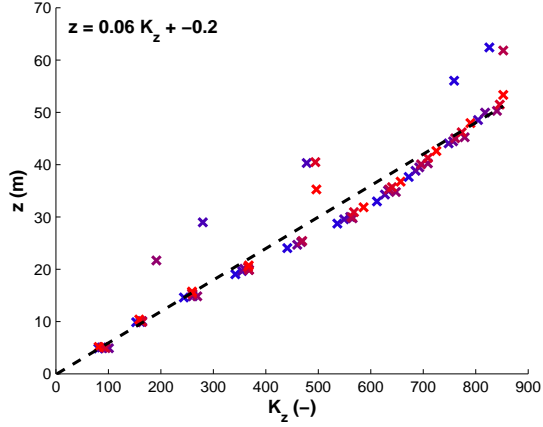} 
\caption{Results of simulated hover tests. The robot is placed at different heights with $K_z=50$ with $\vartheta^{*}=0$. The gain is then adapted with $\mathrm{cov}(u_z', \vartheta_z)^{*}=0.05$. x-markers show the height $z$ and $K_z$ at the first time instant in which $\mathrm{cov}(u_z', \vartheta_z) \geq 0.05$. The markers are color-coded for the wind (blue $=-3$, red $=3$). The black dashed line is a linear fit.} \label{figure:hover_sim} 
\end{figure}

This strategy has been tested in simulation for $z \in \{5, 10, \ldots, 45, 50\}$ and $v_{\mathrm{wind}} \in \{-3, -2, \ldots, 2, 3\}$. The simulation stops when $|e_{\mathrm{cov}}| < 0.005$. Figure \ref{figure:hover_sim} shows for each simulation the last gain and height with a marker. The markers are color-coded for the wind from blue ($-3$ m/s) to red ($3$ m/s). The black dashed line is a linear least squares fit. 

The main result is that this method indeed gives an approximately linear relationship between $z$ and $K_z$, with as linear fit: $z = 0.06 K_z - 0.2$. Although the wind influences the hover altitude (since it is not accounted for in the initial thrust) it again hardly influences the results. Finally, despite the theoretical stability analysis, it may be counter-intuitive that the relation between $K_z$ and $z$ is linear. However, it is instructive to realize that the oscillations at $50$m are higher than at $5$m (in the order of $0.5$m versus $0.1$m). 

\subsection{Distance Estimation During Landing} \label{section:full_monty}
Is there a possibility to land and continuously estimate the distance to the surface? Figure \ref{figure:hover_sim} suggests that there is. If the robot descends while keeping $\mathrm{cov}(u_z', \vartheta_z)$ at a fixed positive value, then the gain $K_z$ will directly represent the height $z$ during the landing. This strategy of ``landing on the edge of oscillation'' can use the exact same adaptive gain control as explained above, but then with $c^2 > 0$. 

\begin{figure}[h!]
\centering
\includegraphics[width=8cm]{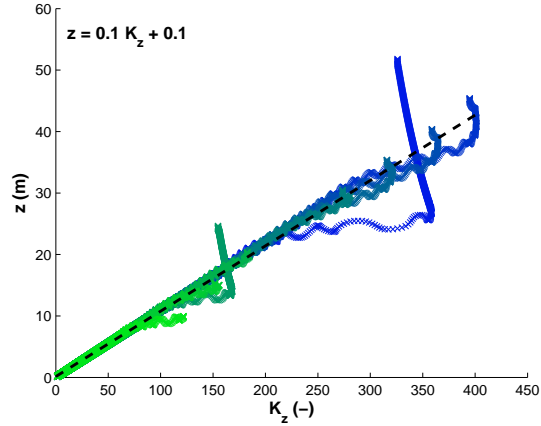}
\caption{Results of simulated tests landing on the edge of oscillation. The robot is placed at different heights with $K_z=50$. Then, it starts adaptive gain control with $\mathrm{cov}(u_z', \vartheta_z)^{*}=0.05$. First $\vartheta^{*}=0$, and when for the first time $\mathrm{cov}(u_z', \vartheta_z) \geq 0.05$, the relative velocity set point is changed to $\vartheta^{*}=0.05$.} \label{figure:landing_on_the_edge}
\end{figure}

Simulations of this strategy have been performed for $z \in {5, 10, \ldots, 50}$. The implementation of the strategy starts with a hover maneuver ($c^2 = 0$) with $K_z=50$, $I = 0.005$, and $P = 0.15$. If in hover the condition $\mathrm{cov}(u_z', \vartheta_z)^{*}=0.2$ is met, the landing is commenced. During landing $c^2=0.05$, $\mathrm{cov}(u_z', \vartheta_z)^{*}=0.05$. 

%

Figure \ref{figure:landing_on_the_edge} shows the $K_z$ versus $z$ during the landing phase. The markers are color-coded for the starting height $z_0$ in the landing phase, from blue ($50$m) to green ($5$m). After switching from $c^2=0$ to $c^2=0.05$, the gain is often too low for causing instability. The outer loop has to compensate for this, which causes $K_z$ to vary at the start of the landing phase. After that, all $K_z$ are linearly related to $z$, with as linear fit: $z = 0.1 K_z + 0.1$. 

\section{Real-world Experiments} \label{section:robot_experiments}

\subsection{Experimental Setup}
\begin{figure}[b!]
\centering
\includegraphics[width=12cm]{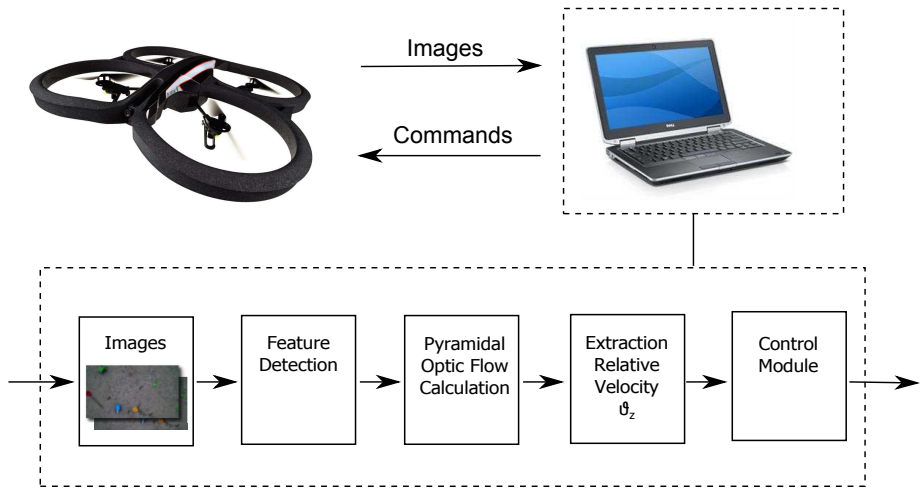}
\caption{Experimental setup. A Parrot AR drone 2.0 sends its images to an offboard laptop, which processes the images and sends back control commands.} \label{figure:setup}
\end{figure}

The experimental setup is shown in Figure \ref{figure:setup}. A Parrot AR drone 2.0 is used, running the Parrot firmware on board. This means that the state estimation and control of attitude and altitude are performed on board, with the help of the drone's sensors. The sensors consist of an IMU (accelerometers, gyros, magnetometer, pressure sensor), and a downward pointing camera and sonar. The firmware can be considered a black box, which takes reference angles and velocities as commands. If no command is received, the drone cancels lateral velocity and keeps its height. 

\begin{figure}[htb!]
\centering
\includegraphics[width=7cm]{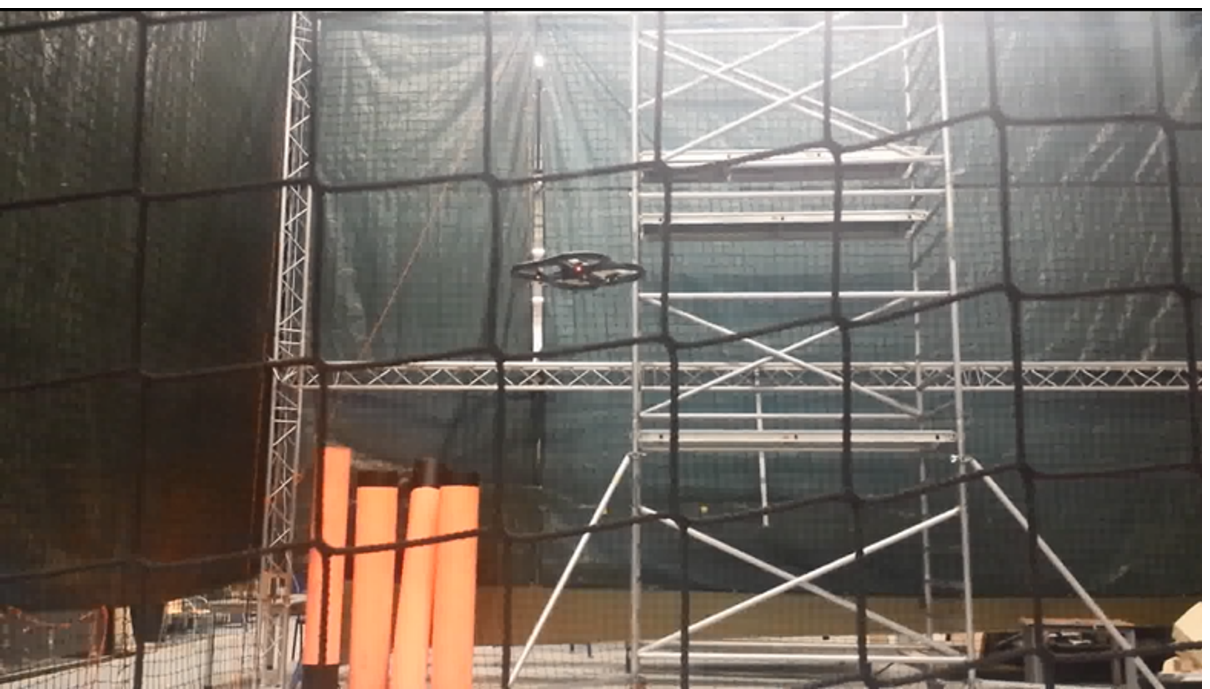}  \includegraphics[width=7cm]{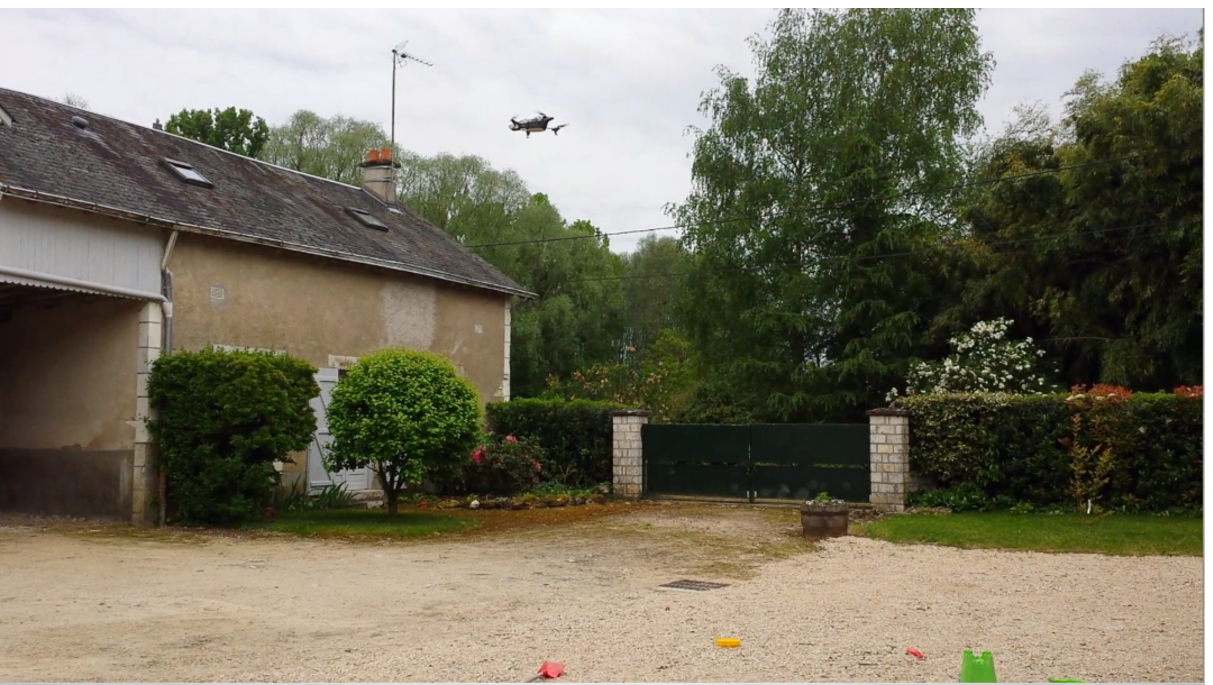}
\caption{\textbf{Left:} Indoor test environment. \textbf{Right:} Outdoor test environment.} \label{figure:environments}
\end{figure}

On the laptop, the in-house developed software package `SmartUAV' is used to communicate with the drone \cite{WAGTER2007}. It receives images from the drone at $\approx15$ Hz. These images are processed with a rather standard vision pipeline. Maximally 40 corners are detected with the method in \cite{SHI1994}, and these corners are tracked to the next image with the Lucas-Kanade optical flow algorithm \cite{LUCAS1981}. The method from \cite{DECROON2013} is used to estimate $\vartheta_z$ from the optical flow vectors. Second order terms are omitted, as was done in \cite{ALKOWATLY2014}. It assumes a flat landing surface and estimates the spatial gradient of the optical flow for determining $\hat{\vartheta_z}$. This value is low-pass filtered and then used in a control loop $v_z' = K_z (\vartheta_z^{*} - \hat{\vartheta_z})$, where $v_z'$ is the vertical velocity command sent to the Parrot AR drone. Please note that in the real-world experiments, the velocity command is saturated at $|v_z'| = 0.75$m/s. The program further allows to adapt $K_z$ according to the adaptive gain control strategy explained in Section \ref{section:full_monty}. Please note that the offboard vision processing setup introduces a significant delay between the capturing of the image and the reception of a command signal from the offboard laptop, of $\sim1$-$1.5$s. Experiments were performed in an indoor and an outdoor environment, shown in Figure \ref{figure:environments}.

Before showing the results of the various tests, it is worthwhile reflecting on the difference between the real-world experimental setup and the simulation setup. Most importantly, the main issue at stake here is whether $K_z$ contains information on the height $z$, by performing optical flow control on the basis of the relative vertical velocity $\vartheta_z$. A major difference with the experiments in simulation is that a velocity $v_z'$ is controlled instead of $u_z'$ and that the firmware uses sensors such as sonar to perform this velocity control. This may lead to the impression that scaling information is still necessary. However, the value of $v_z'$ is only used for determining a covariance (whether the control loop wants to go faster up or down), and its value in meters per second does not play a role in the estimation. Moreover, in the simulations $u_z'$ was controlled directly, demonstrating that the use of $v_z'$ is not necessary to make the approach work. The experiments verify whether the results from theory and simulation also apply to a real-world system with significantly different actuation dynamics.

%
%
\subsection{Indoor Experiments} \label{section:indoor}
\begin{figure}[h!]
\centering
\includegraphics[width=10cm,trim=0mm 2mm 0mm 2mm, clip]{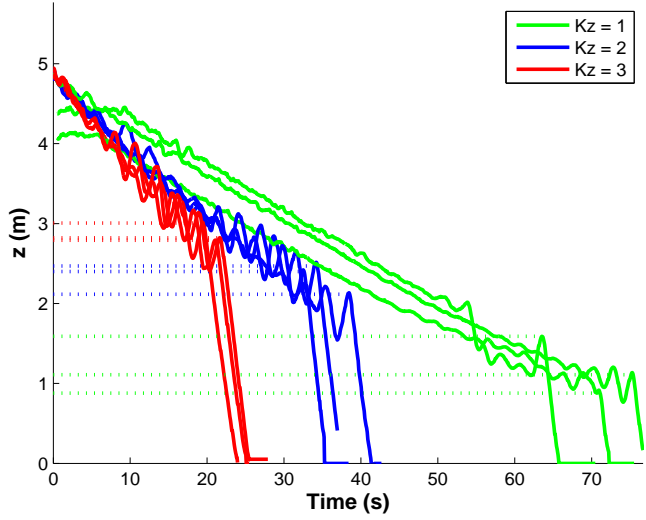} 
\caption{Three fixed-gain constant divergence landings of a Parrot AR drone, with set point $\vartheta_z^{*}=-0.1$ and gains $K_z \in \{ 1, 2, 3 \}$ (green solid, blue dashed-dotted, and red dashed lines, respectively). The dotted lines show the point at which $\mathrm{cov}(u_z', \vartheta_z) = 0.1$ indicating the onset of self-induced oscillations. Given a fixed gain, this instability depends on the height.} \label{figure:fixed_gain_landing}
\end{figure}

First, it is verified that a fixed-gain constant divergence landing results in instability at different heights. The result of this test for $K_z \in \{1, 2, 3\}$ (green, blue, and red lines) and $c^2=0.1$ is shown in Figure \ref{figure:fixed_gain_landing}. Dotted lines indicate the first point at which $\mathrm{cov}(u_z', \vartheta_z) = 0.1$, set to trigger the drone's automatic landing procedure. The main observation from the figure is that also for the real robot, higher gains lead to an instability at larger heights. 

\begin{figure}
\centering
\includegraphics[width=6cm]{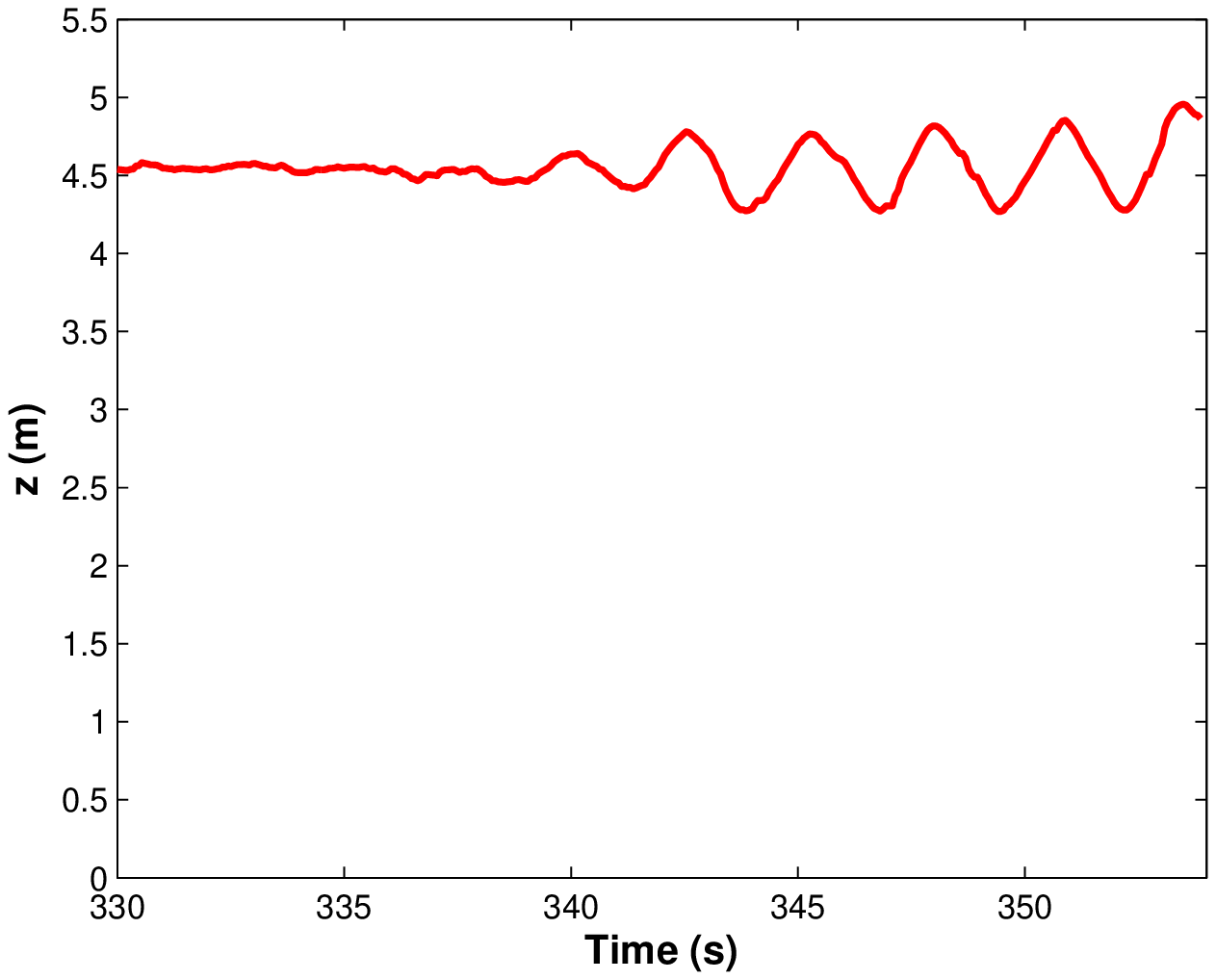} \includegraphics[width=6cm]{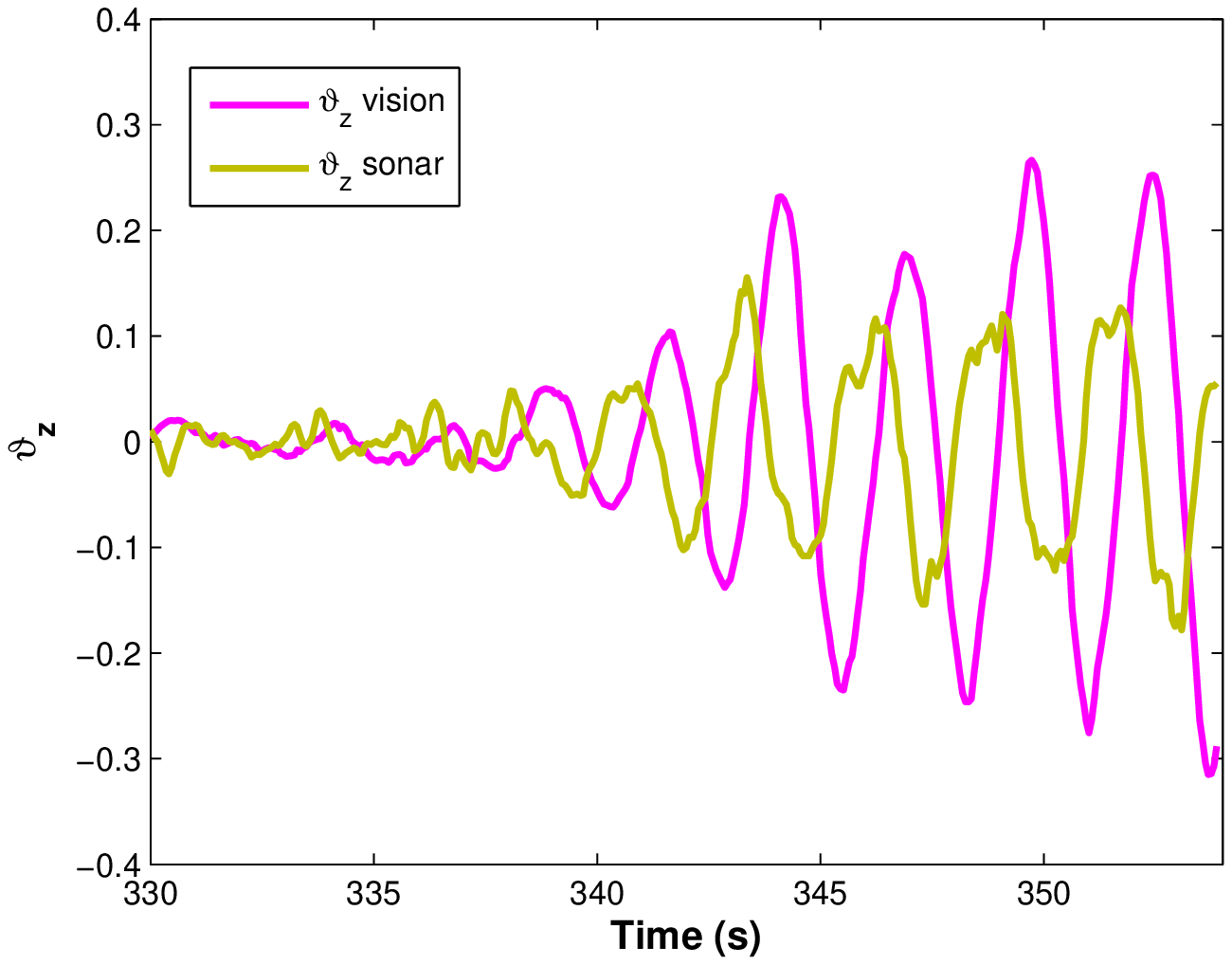}
\includegraphics[width=6cm]{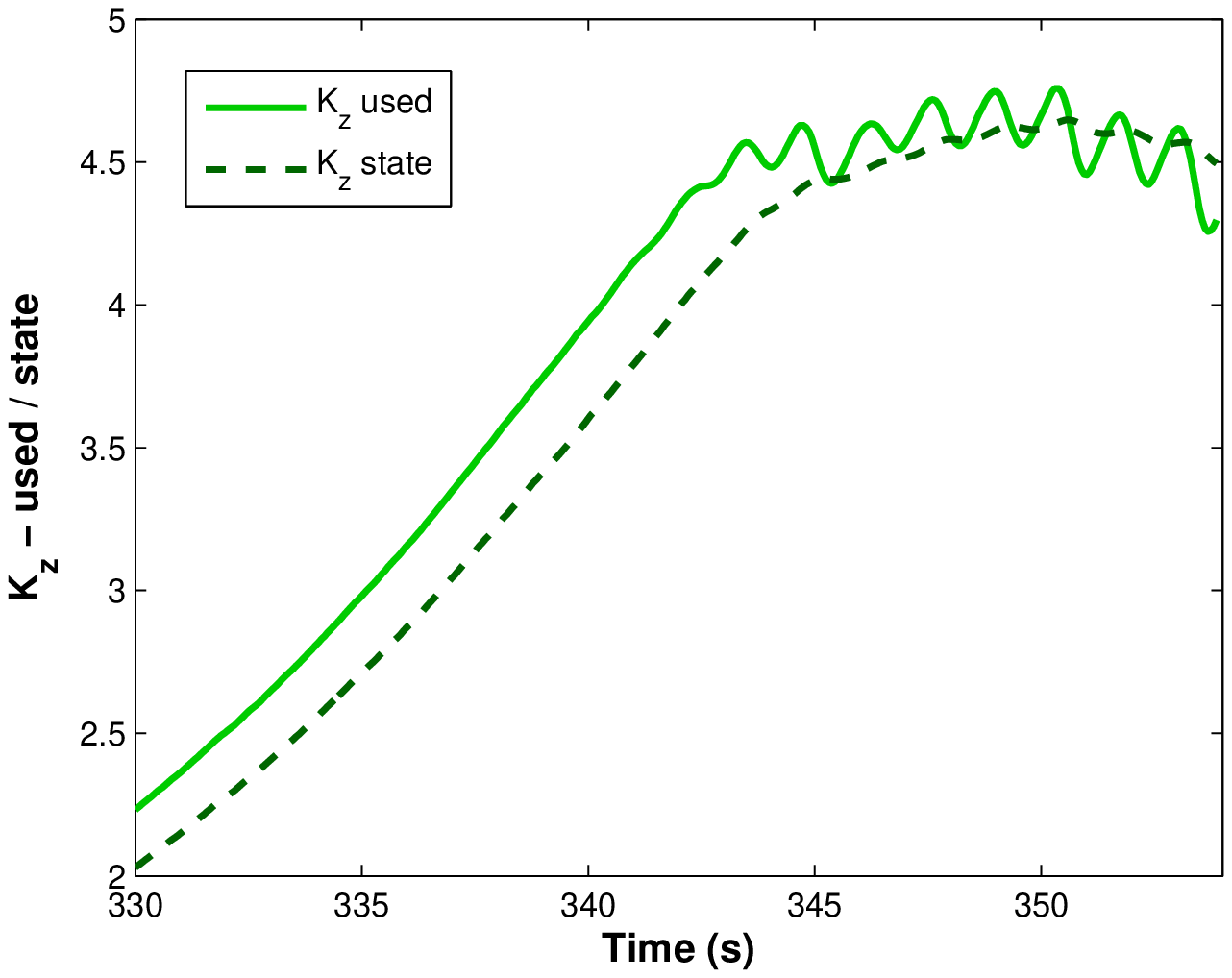} \includegraphics[width=6cm]{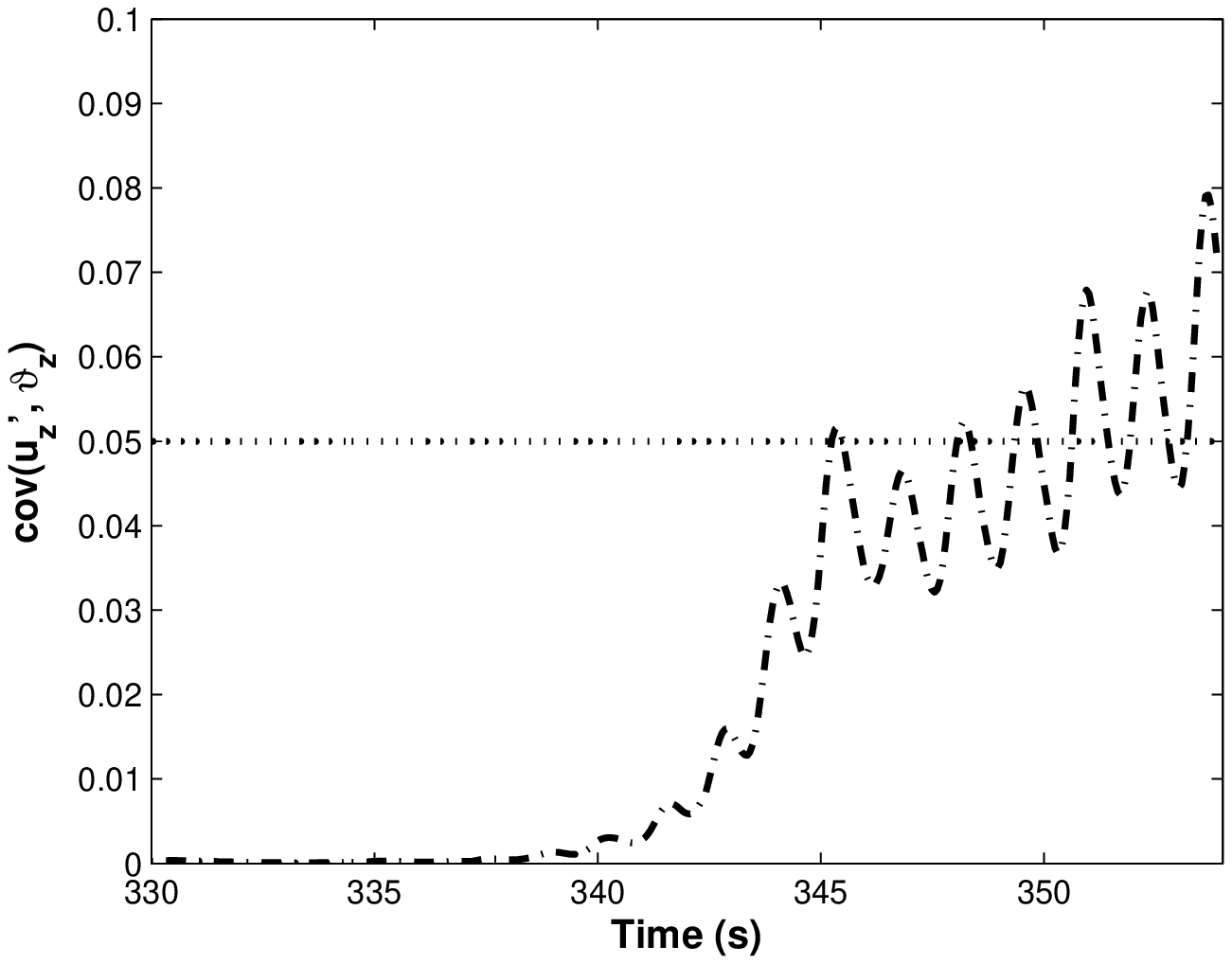}
\caption{Signals during a part of the indoor hover experiment. \textbf{Top left:} The height over time. \textbf{Top right:} Relative velocity over time. The magenta line is the $\vartheta_z$ used for control, as determined with computer vision, the yellow line is the $\vartheta_z$ as calculated with the help of the sonar readings (not used in control). \textbf{Bottom left:} The gain over time. The light green solid line is $K_z$, the dark green dashed line is $K_z'$ (see Eq. \ref{PI-control-Kz} and \ref{PI-control-Kz-prime}). \textbf{Bottom right:} The dashed-dotted line is the covariance of the control input $u_z'$ and $\vartheta_z$, the dotted line is the set point for the covariance ($\mathrm{cov}(u_z', \vartheta_z)^{*}$, see Eq. \ref{adaptive_error}).} \label{figure:hover_zoom_indoors}
\end{figure} 

\begin{figure}
\centering
\includegraphics[width=7cm]{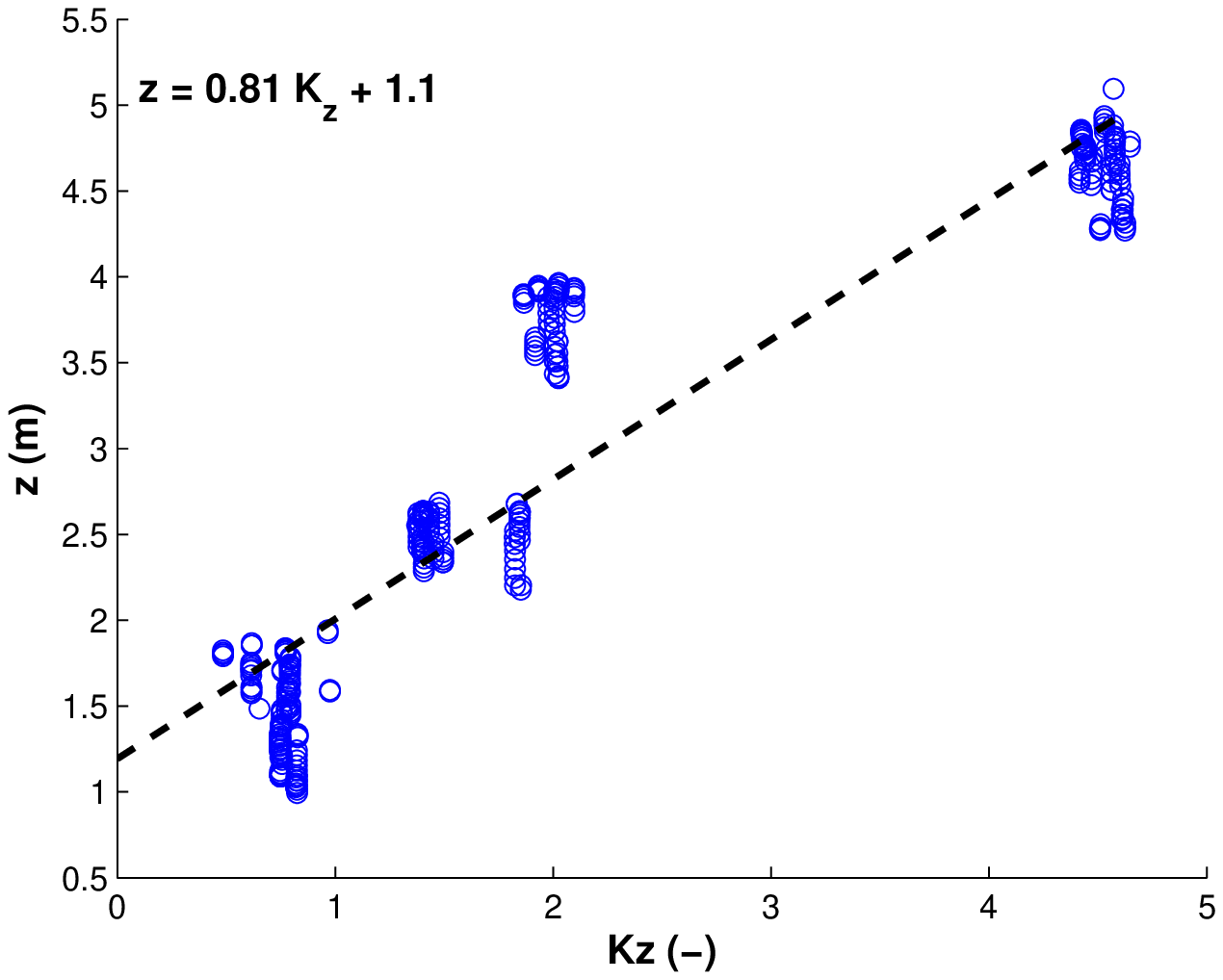} \includegraphics[width=7cm]{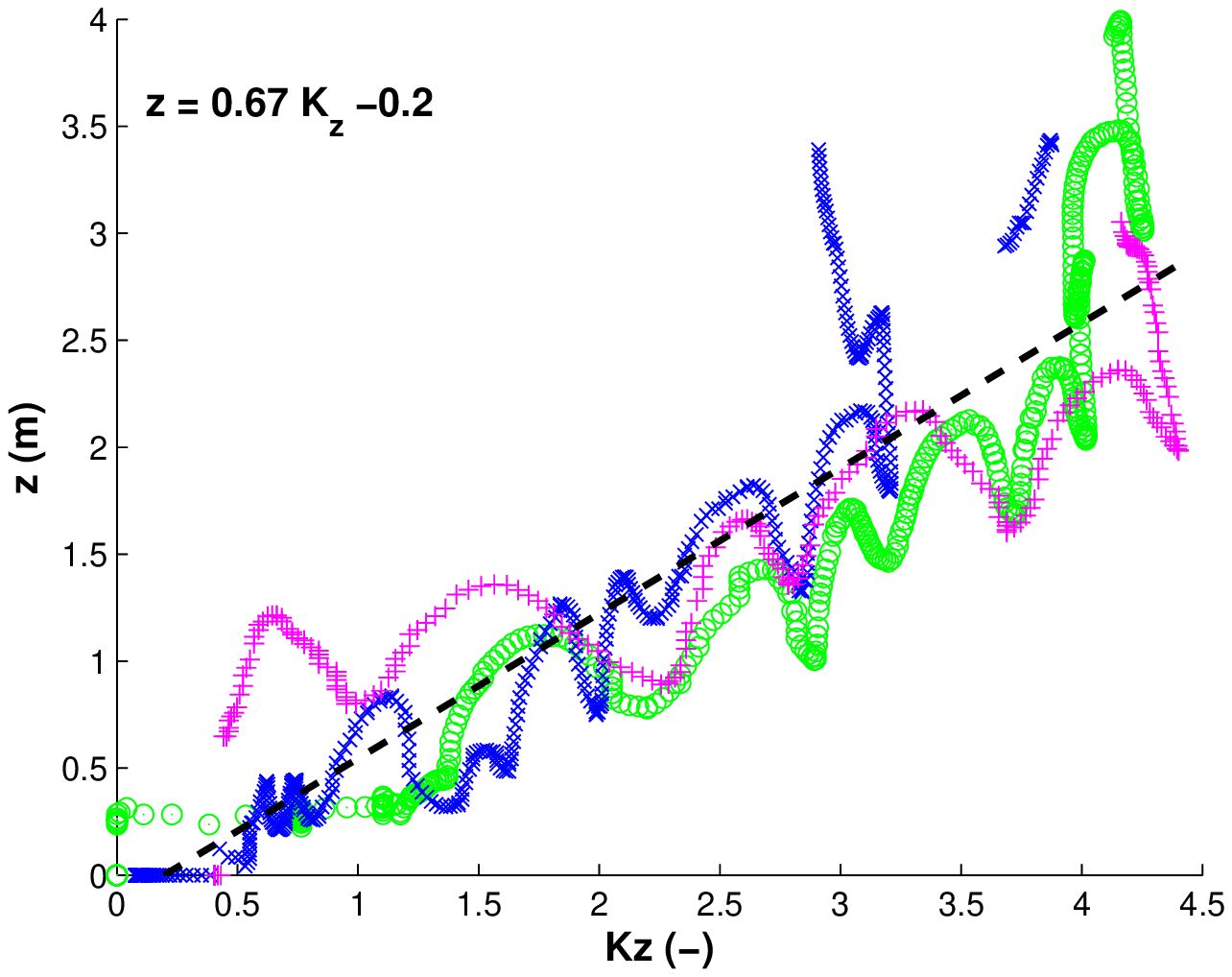} 
\caption{Indoor results adaptive gain control. \textbf{Left:} Hover experiments at different altitudes. The markers represent $(K_z, z)$ values for which $|e_{\mathrm{cov}}| < 0.005$. \textbf{Right:} Landing on the edge of oscillation. The markers are $(K_z, z)$-pairs during the landing, while $K_z$ is regulated so that $\mathrm{cov}(u_z', \vartheta_z) = 0.05$. Markers are only shown at instants at which $|e_{\mathrm{cov}}| < 0.005$.} \label{figure:hover_real}
\end{figure}

The second finding to verify is the determination of height around hover. For the real robot, $\mathrm{cov}(u_z', \vartheta_z)^{*} = 0.05$ is employed, with $I = 0.075$ and $P = 2$. During the hover experiment, the robot has been sent to different heights, subsequently activating the adaptive gain control. In order to get an impression of what the adaptive control does, let us first zoom in on the relevant variables during a short part of the flight, shown in Figure \ref{figure:hover_zoom_indoors}. In this part of the flight, the drone has just ascended to $\sim4.5$m (see top left plot), while it has a gain of $K_z \approx 2.2$ (bottom left). The drone starts to regulate its relative velocity to $\vartheta_z^{*} = 0$ (top right), and initially there are no self-induced oscillations as witnessed by the $\mathrm{cov}(u_z', \vartheta_z) \approx 0$  (bottom right). The low covariance makes the gain $K_z'$ increase over time. This gradually causes $\vartheta_z$ to show ever larger oscillations, which leads to higher  $\mathrm{cov}(u_z', \vartheta_z)$. Around $t = 174$s, the covariance reaches its set point, and the gain $K_z'$ starts stabilizing around $4.6$ in order to keep the covariance around $0.05$. The question now is at what values $K_z'$ will stabilize at other heights, and whether there is indeed a linear relationship between $z$ and $K_z$.


The left part of Figure \ref{figure:hover_real} shows the result of having the drone hover at a given height with $\mathrm{cov}(u_z', \vartheta_z)^{*} = 0.05$. The markers are only shown for time instants at which $|e_{\mathrm{cov}}| < 0.005$. The dashed line is a linear least squares fit. Although more noisy than in simulation, the relation between $z$ and $K_z$ seems to be roughly linear - as predicted by the theoretical derivation. When using the linear fit $\hat{z} = 0.81 K_z + 1.1$ as a height estimation function, the median absolute error of all observations is $|z-\hat{z}| = 0.28$m, while the mean is $0.40$m.%

\begin{figure}[htb]
\centering
\includegraphics[width=7cm]{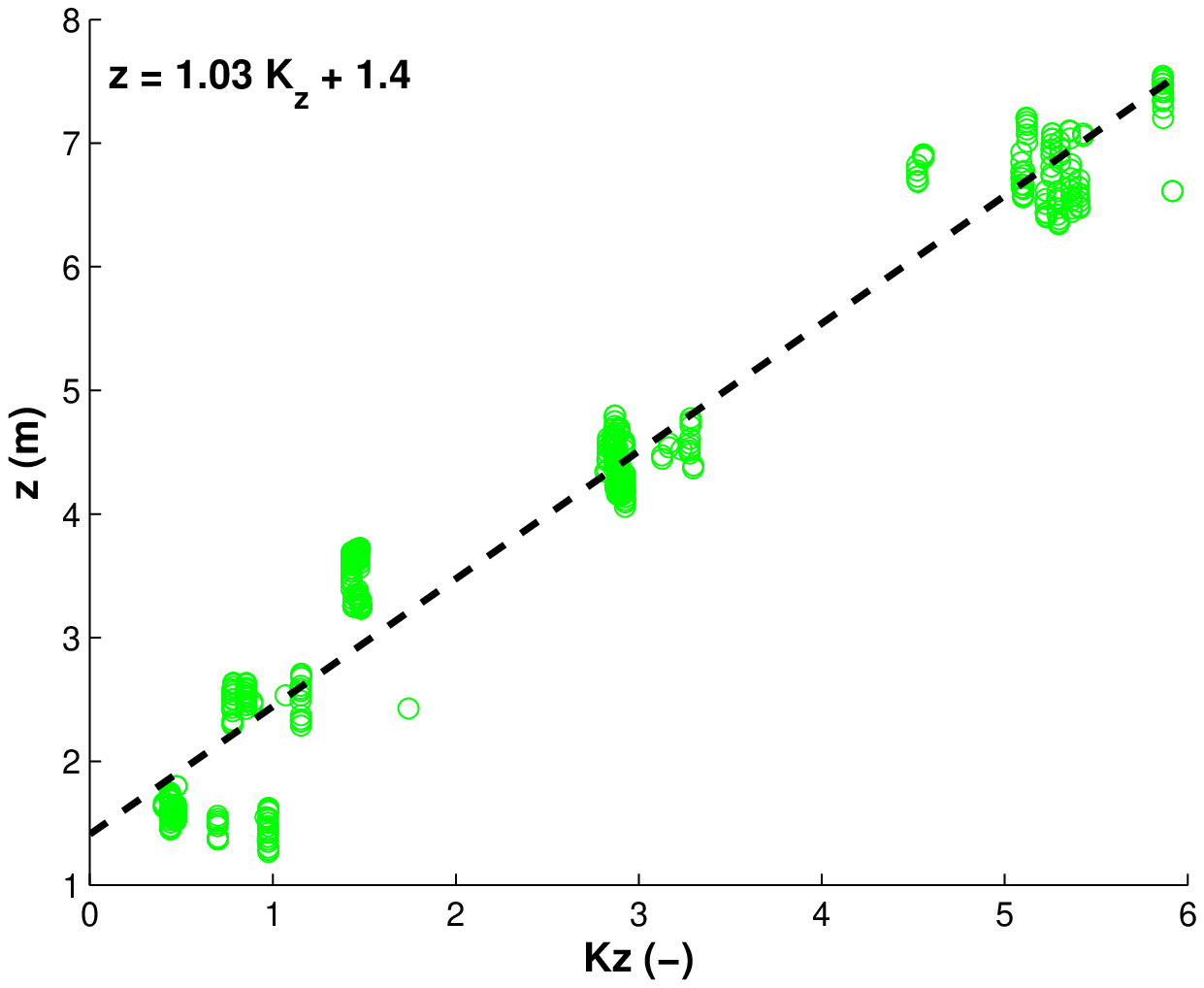} \includegraphics[width=7cm]{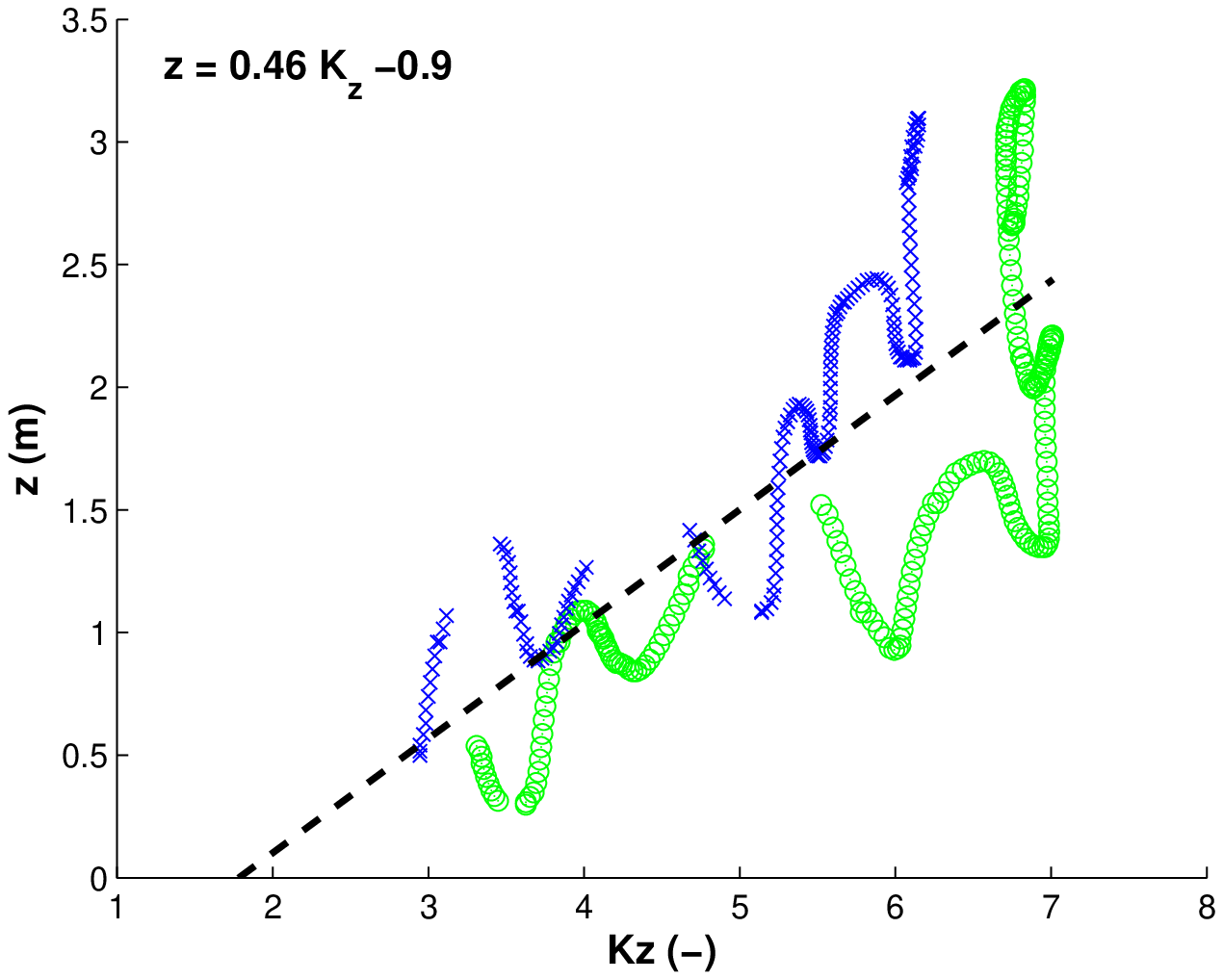}
\caption{Outdoor results adaptive gain control. \textbf{Left:} Results of hover experiments at different altitudes, with $\mathrm{cov}(u_z', \vartheta_z)^{*} = 0.05$. The markers represent $(K_z, z)$ values for which $\mathrm{cov}(u_z', \vartheta_z) \in [0.005, 0.015]$. \textbf{Right:} Results of landing on the edge of oscillation. The markers are $(K_z, z)$-pairs during the landing, while $K_z$ is regulated so that $\mathrm{cov}(u_z', \vartheta_z) = 0.005$.} \label{figure:outdoor_exps}
\end{figure}

The third finding to verify is the possibility to land on the edge of oscillation. The AR drone is sent to a height of $3.5$m. Then, the adaptive gain control is started, at first with $c^2=0$. As soon as $|e_{\mathrm{cov}}| < 0.005$, the landing phase starts. If $K_z \leq 0.5$, the drone's automatic landing procedure is triggered. The right part of Figure \ref{figure:hover_real} shows the results for all points where $|e_{\mathrm{cov}}| < 0.005$. Different experiments have different markers and colors. The dashed line is a linear fit. Although the results are quite noisy, again a roughly linear relation appears, with as linear fit $z = 0.67 K_z - 0.2$. The median absolute error of this fit is $0.26$m, and the mean absolute error is $0.33$m.  

\subsection{Outdoor Experiments} \label{section:outdoor}
A few additional outdoor experiments have been performed to confirm that the method also works in an outdoor environment with wind and wind gusts. The right part of Figure \ref{figure:environments} shows the outdoor experimental setup. The ground surface consisted of small stones, which already provide ample texture. Some beach toys were added to the scene to also ensure visible texture at larger heights. The left part of Figure \ref{figure:outdoor_exps} shows the results from the hover experiment. Just as indoors, there is a clear linear relationship between the height $z$ as measured by the sonar and the gain $K_z$ as determined with the adaptive gain control, with as linear fit $z = 1.03 K_z + 1.4$. The median absolute error in height of the fit is $0.29$m, the mean is $0.35$m.  The right part of Figure \ref{figure:outdoor_exps} shows the results from two landings on the edge of oscillation, again showing a linear relationship. Videos of some of the experiments can be seen at\footnote{\url{https://www.youtube.com/playlist?list=PL_KSX9GOn2P96xb8JRSo0jLFnLBK0Pj5Y}}. 




\section{Discussion} \label{section:discussion}

\subsection{Flying insects}

The presented strategy forms a novel hypothesis on how flying insects such as fruitflies and honeybees can estimate distances, viz. by means of the detection of self-induced oscillations. In \cite{BREUGEL2014} a different, thrust-based strategy for distance estimation was introduced, and its implications for landing insects discussed. In part, the implications of these strategies overlap. They both explain how flying insects can trigger landing responses at a particular given distance, such as fruitflies that extend their legs just before landing (e.g., \cite{BREUGEL2012}). However, there is a significant difference in the strategies, which has its consequences for the explanation of flying insects' behavior. A few notable differences in this respect will be highlighted below. 


A first difference is in the possible explanation of the hover phase honeybees exhibit just before touchdown \cite{EVANGELISTA2010}. The hover phase could in principle be explained as a maneuver that is triggered by a distance measurement - for instance by means of the thrust-based strategy of \cite{BREUGEL2014}. However, the proposed strategy suggests instead that the hover mode is an intrinsic property of optical flow control and is part of the distance estimation itself. As the honeybee gets closer to the surface, its control gain will start becoming too high. This will result in self-induced oscillations and lead to a situation of zero divergence, i.e., hover. In this light, it is interesting to remark the characteristic little `bump' in the trajectory in Fig. \ref{figure:detection}. The same bump can be observed in Fig. 4 in \cite{EVANGELISTA2010}, and in Fig. 5 in \cite{SRINIVASAN1996} (in which a honeybee is performing a grazing landing). 


A second difference is that the stability-based strategy explicitly allows for the observation that a visual stimulus itself can trigger the landing responses - even for tethered insects \cite{borst1986kind,TAMMERO2002}. The reason for this is that the strategy depends on the control stability and hence the system bandwidth. A standard way to measure the system bandwidth is to make a Bode plot that maps the system's frequency response. At too high frequencies, the phase shift between the quickly changing observation and the subsequent control input will become larger and the magnitude of the control input lower. If the magnitude of the control input is large enough, a purely visual stimulus may be detected as a self-induced oscillation. 

%
%
%
%
%
%
%
%

\subsection{Flying robots}
Concerning flying robots, the novel proposed distance estimation strategy provides a novel way to perceive distances with a single vision sensor. This may turn out to be essential for tiny flying robots such as the Robobee \cite{MA2013}, for which any type of sensor is a significant payload \cite{imav2014:s_fuller_et_al,FULLER2014,HELBLING2014}. For slightly larger drones such as 25g `pocket drones' (e.g., \cite{DUNKLEY2014}), the finding is also immediately relevant. For instance, currently the lightest, fully autonomously flying robot is the 20g DelFly Explorer \cite{DEWAGTER2014}, which uses a $4$g stereo vision system with onboard processing to avoid obstacles in its environment. The proposed stability-based strategy to distance estimation may provide a bigger potential to even lighter, single-camera solutions, also for obstacle avoidance. Even for MAVs in the order of $\sim1$kg or larger, the method may still be useful. It allows to estimate potentially large distances without compromising the MAVs' payload capability. %

However, in order for stability-based distance estimation to live up to its potential, some issues need to be further investigated. The experiments have shown the strategy to work in an offboard processing scheme with a Parrot AR drone $2.0$ with a median height estimation error in the order of $\sim0.3$m. As a proof of concept this suffices, but it is desirable to improve the accuracy of the measurement. This seems quite possible if onboard processing is used, so that higher update frequencies can be achieved (e.g., $60$Hz with onboard processing of the bottom camera images on the AR drone $2.0$).  


\section{Conclusion} \label{section:conclusion}
In this paper, a stability-based strategy has been proposed with which a robot can estimate distances only on the basis of efference copies and optical flow maneuvers. Theoretical analysis of linearized models has shown that there is a linear relation between the fixed gain and the height at which instability arises. This analysis has been verified in simulation (including non-linearities and disturbing factors such as wind gusts) and on a real robotic platform. It has been demonstrated that self-induced oscillations can be detected by the robot and used to (1) trigger a final landing procedure, (2) determine the height in hover, and (3) continuously determine the height during a constant divergence landing.

\section*{Acknowledgements}
I would like to thank Christophe De Wagter, Hann Woei Ho, and Tobias Seidl for the interesting discussions and comments on earlier versions of the manuscript.

\bibliographystyle{plain}
\bibliography{bibliogdc}

\appendix
\section{Appendix: Generalization to other movement directions} \label{section:other_movement_directions}
In Section \ref{section:distance_estimation} it was shown that given a gain $K_z$, the stability of the vertical control loop with $\vartheta_z$ depends on the height $z$. In this appendix the case is studied in which the camera still looks down in the $z$-direction, but the movement is in the horizontal $x$-direction. The control is thus based on the relative velocity $\vartheta_x = \frac{v_x}{z}$. It will first be shown that the finding also generalizes to such horizontal optical flow control (Subsection \ref{section:stab_app}). The stability-based method then does not require vertical motion. Subsequently, a thrust-based method is investigated for horizontal motion (Subsection \ref{section:thrust_app}). The thrust-based method always requires both horizontal and vertical motion. 

\subsection{Stability-based Method} \label{section:stab_app}%
Let us start from the observation:
\begin{equation}
y = \vartheta_x = \frac{v_x}{z},
\end{equation}
\noindent Since both the horizontal and vertical axes are involved, the state space model will have four variables. Linearizing the state space model gives:
\begin{equation}
\Delta y(t) = \begin{bmatrix} \frac{-v_x}{z^2} & 0 & 0 & \frac{1}{z} \end{bmatrix} \begin{bmatrix} \Delta z(t) & \Delta v_z(t) & \Delta x(t) & \Delta v_x(t) \end{bmatrix}^{\top},
\end{equation}
\noindent so that the state space model matrices are:
\begin{equation} \label{ss_continuous_app}
A = \begin{bmatrix} 0 & 1 & 0 & 0\\ 0 & 0 & 0 & 0\\ 0 & 0 & 0 & 1\\ 0 & 0 & 0 & 0 \end{bmatrix} \: \: B = \begin{bmatrix} 0 \\ 0\\ 0\\ 1 \end{bmatrix} \\ 
C = \begin{bmatrix} \frac{-v_x}{z^2} &  0 & 0 & \frac{1}{z} \end{bmatrix} \: \: D = [0].
\end{equation}

Again, the discretized system is studied, which has the following state space model matrices corresponding to the continuous ones in Eq. \ref{ss_continuous_app}:
\begin{equation} \label{ss_discrete_app}
\Phi = \begin{bmatrix} 1 & T & 0 & 0\\ 0 & 1 & 0 & 0\\ 0 & 0 & 1 & T\\ 0 & 0 & 0 & 1 \end{bmatrix} \: \: \Gamma = \begin{bmatrix} 0\\ 0 \\ \frac{T^2}{2}\\ T  \end{bmatrix} \\
C = \begin{bmatrix} \frac{-v_z}{z^2} & 0 & 0 & \frac{1}{z} \end{bmatrix} \: \: D = [0],
\end{equation} 
\noindent where $T$ is the discrete time step in seconds. The transfer function of the open loop system can be determined to be:
\begin{equation}
G(w) = C (w I - \Phi)^{-1} \Gamma,
\end{equation}
\begin{equation}
= \frac{T}{z(w-1)},
\end{equation} 
\noindent where we use $w$ as the $Z$-transform variable, since $z$ already represents height. The feedback transfer function is:
\begin{equation} \label{general_transfer_function_app} 
G(w) = \frac{K_x T}{z (w-1) \left( \frac{K_x T}{z (w-1)} + 1 \right)}.
\end{equation}
Eq. \ref{general_transfer_function_app} shows that, given a gain $K_x$ and time step $T$, the dynamics and stability of the system depend on the height $z$. 

Setting $w=-1$ in the denominator and equating it with $0$ gives:
\begin{equation}
-2z \left( \frac{K_x T}{-2z} + 1 \right) = 0
\end{equation}
\noindent , which leads to an equation that expresses the height at which the control system will become unstable in terms of the unstable gain value $K_x$:
\begin{equation} \label{vacuum_z_unstable_K_app}
z = \frac{1}{2} K_x T,
\end{equation}
\noindent which is exactly the same formula as for the vertical movements studied in Section \ref{section:distance_estimation}.

\subsection{Thrust-based Method} \label{section:thrust_app}
It is instructive to also look at what a possible thrust-based method would look like for the horizontal motion direction. For this, let us again start from the observation $\vartheta_x$, which is differentiated over time:
\begin{equation} 
\dot{\vartheta_x} = \frac{a_x}{z} - \frac{v_x v_z}{z^2} = \frac{a_x}{z} - \vartheta_x \vartheta_z,
\end{equation}
\noindent leading to:
\begin{equation} \label{z_hor}
z = \frac{a_x}{( \dot{\vartheta_x} + \vartheta_x \vartheta_z )}.
\end{equation}
\noindent Eq. \ref{z_hor} shows that the height can be estimated with the help of a horizontal accelerometer measurement and the observables $\vartheta_z$, $\vartheta_x$, and the time derivative $\dot{\vartheta_x}$ (as was done in \cite{IZZO2013}). 

If the relative velocity $\vartheta_x$ (sometimes referred to as `ventral flow') is kept constant, then $\dot{\vartheta_x} = 0$, and:
\begin{equation} \label{z_hor_simple}
z = \frac{a_x}{\vartheta_x \vartheta_z},
\end{equation}
\noindent where in a vacuum environment, $a_x$ can be assumed to be $u_x$. A constant ventral flow and divergence landing would then again lead to a way to estimate $z$, as $u_x$ is known and $\vartheta_x$ and $\vartheta_z$ can be assumed equal to constants.

Interestingly, also this horizontal motion case does not allow for distance estimation while staying at the same height ($\vartheta_z = 0$). The reason for this can be seen when rearranging Eq. \ref{z_hor_simple}:
\begin{equation} \label{ax_hor}
a_x = z (\vartheta_x \vartheta_z),
\end{equation}
\noindent which shows that $\vartheta_z=0$ implies $a_x=0$. Indeed, if $\vartheta_z=0$, $\vartheta_x$ can only be constant if there is no horizontal acceleration. 

\end{document}